\documentclass{article} %
\usepackage{colm2024_conference}

\usepackage{booktabs}
\usepackage{graphicx}
\usepackage{enumitem}
\usepackage{wrapfig}
\usepackage[raster,skins]{tcolorbox}
\usepackage{algorithm}
\usepackage{algpseudocode}
\usepackage{subcaption}
\usepackage{microtype}
\usepackage{amsmath}
\usepackage{colortbl}
\usepackage[utf8]{inputenc}
\definecolor{lightgray}{rgb}{0.9,0.9,0.9}
\usepackage{caption}
\usepackage{subcaption}
\usepackage{xcolor}
\usepackage{setspace}
\usepackage{url}
\usepackage{multirow}
\usepackage{colortbl}
\usepackage{tabularx}
\usepackage{blindtext}
\usepackage{pgfplots}
\pgfplotsset{compat=1.18} 
\usepackage{tikz}
\usetikzlibrary{er,positioning,bayesnet}
\usepackage{makecell}
\usepackage{tipa}
\usepackage{siunitx}
\usepackage{nicefrac}
\usepackage{tocloft}
\usepackage{listings}
\usepackage[raster,skins]{tcolorbox} %
\usepackage{xltabular}
\usepackage{adjustbox}
\usepackage{xurl}
\usepackage{booktabs} % 高质量表格线
\usepackage{array}    % 自定义列格式
\usepackage{caption}  % 美化 caption
\usepackage{tabularx}
\usepackage{amssymb}
\usepackage{tocloft}

\usepackage{amsmath,amsfonts,bm}

\def\eqref#1{equation~\ref{#1}}
\def\1{\bm{1}}

\DeclareMathAlphabet{\mathsfit}{\encodingdefault}{\sfdefault}{m}{sl}
\SetMathAlphabet{\mathsfit}{bold}{\encodingdefault}{\sfdefault}{bx}{n}

\newcommand{\modelname}{JT-DA}

\title{\modelname{}: Enhancing Data Analysis with Tool-Integrated Table Reasoning Large Language Models}

\author{
\bf Ce Chi\textsuperscript{\rm 1 \dag}, Xing Wang\textsuperscript{\rm 1 \dag \rm *}, Zhendong Wang\textsuperscript{\rm 1 \dag}, Xiaofan Liu\textsuperscript{\rm 1 \dag}, Ce Li\textsuperscript{\rm 1 \dag}, \\Zhiyan Song\textsuperscript{\rm 1 \dag}, Chen Zhao\textsuperscript{\rm 1 \dag}, Kexin Yang\textsuperscript{\rm 1 \dag}, Boshen Shi\textsuperscript{\rm 1 \dag}, Jingjing Yang\textsuperscript{\rm 1 \dag}, \\Chao Deng\textsuperscript{\rm 1}, and Junlan Feng\textsuperscript{\rm 1 \rm *}
\\\vspace{0.5cm}
\textsuperscript{\rm 1}Jiutian Research, China Mobile, Beijing, China
}

\footnotetext{\rm *Corresponding author: \{wangxing, fengjunlan\}@cmjt.chinamobile.com}
\footnotetext{\rm \dag Equal contribution}

\begin{document}

\maketitle

\begin{abstract}
In this work, we present JT-DA-8B (JiuTian Data Analyst 8B), a specialized large language model designed for complex table reasoning tasks across diverse real-world scenarios. To address the lack of high-quality supervision in tabular reasoning scenarios, we construct a comprehensive and diverse training corpus with 34 well-defined table reasoning tasks, by aggregating 29 public table QA datasets and 3 million tables. An automatic pipeline is proposed to generate realistic multi-step analytical tasks involving reasoning patterns. The model is trained upon open-source JT-Coder-8B model, an 8B-parameter decoder-only foundation model trained from scratch. In the training stage, we leverage LLM-based scoring and workflow-aligned filtering to distill high-quality, table-centric data. Both supervised fine-tuning (SFT) and Reinforcement learning (RL) are adopted to optimize our model. Afterwards, a four-stage table reasoning workflow is proposed, including table preprocessing, table sensing, tool-integrated reasoning, and prompt engineering, to improve model interpretability and execution accuracy. Experimental results show that JT-DA-8B achieves strong performance in various table reasoning tasks, demonstrating the effectiveness of data-centric generation and workflow-driven optimization.

% \begin{figure*}[htbp]
% \centering
% \begin{subfigure}[b]{0.48\textwidth}
%   \centering
%   \includegraphics[width=\textwidth]{figures/benchmark2.png}
%   \caption{benchmark}
%   \label{fig:benchmark1}
% \end{subfigure}
% \hfill
% % \begin{subfigure}[b]{0.48\textwidth}
% %   \centering
% %   \includegraphics[width=\textwidth]{figures/benchmark3.png}
% %   \caption{benchmark}
% %   \label{fig:benchmark2}
% % \end{subfigure}
% \caption{benchmark}
% \label{fig:benchmark}
% \end{figure*}

\begin{figure*}[htbp]
\centerline{\includegraphics[width=1.0\textwidth]{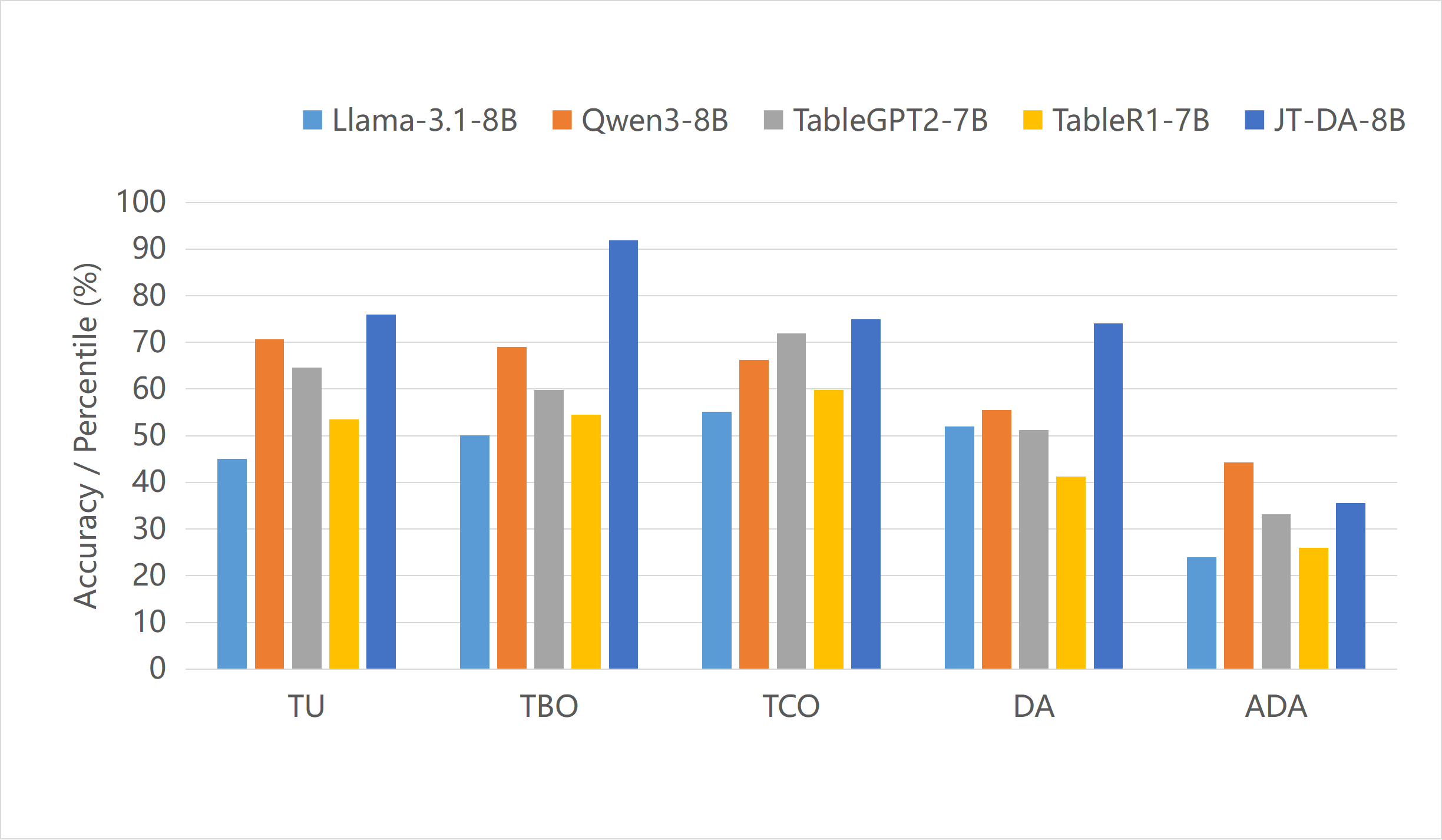}}
\caption{Benchmark performance of JT-DA-8B.}
\label{fig:benchmark}
\end{figure*}

\end{abstract}
\clearpage
% \begingroup
% \renewcommand\contentsname{Contents} % 目录标题
\tableofcontents
% \endgroup
\clearpage

\section{Introduction}
Large language models (LLMs) have transformed the landscape of natural language processing and code generation. Models such as GPT, Gemini \cite{comanici2025gemini} and their open-source counterparts \cite{dubey2024llama,qwen3technicalreport} have achieved human-level performance on a wide range of tasks, including text generation, translation, code completion, and conversational agents, leveraging massive pre-training on diverse corpora of natural language and programming data. 
Recently, the deep reasoning and chain-of-thought capabilities of LLMs have further enabled significant breakthroughs in complex problem-solving across research and industry \cite{deepseekr1}.

While LLMs, including code generation models, have demonstrated impressive capabilities in text-based tasks, table-based reasoning remains a significantly under-explored frontier. 
First, code-oriented LLMs often lack structural awareness when operating on tables \cite{su2024tablegpt2}. 
Frequent errors occur in code-oriented LLMs such as hallucinated or incorrect column names, improper use of schema fields, and confusion between rows and headers. 
These failures arise because existing models are typically trained on generic programming or text data with little or no exposure to structured tabular formats. 
As a result, the models fail to ground their reasoning steps in the actual structure of the table.
Second, while recent works have begun to fine-tune LLMs specifically for table reasoning tasks \cite{yang2025table}, they generally do not incorporate tool integration mechanisms. 
Consequently, these models rely entirely on in-model computation, which often leads to inaccurate results on tasks requiring symbolic manipulation—such as arithmetic, logical comparisons, or aggregation. 
Without tool augmentation, the model must ``guess'' the intermediate analysis results of complicated data analysis calculation that would otherwise be precisely handled by symbolic tools (e.g., calculators, code sandboxes), limiting both performance and reliability \cite{yang2025code}.

% 加个case分析
\begin{figure*}[htbp]
\centerline{\includegraphics[width=1.0\textwidth]{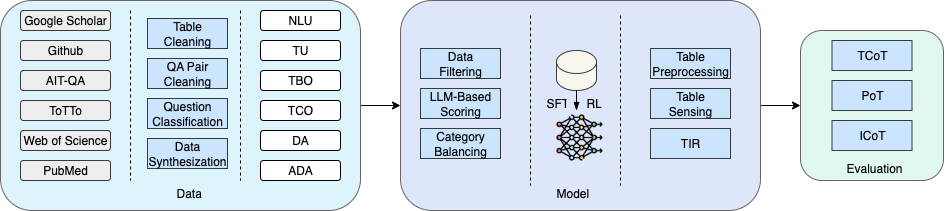}}
\caption{Overall JT-DA-8B architecture.}
\label{fig:overall}
\end{figure*}

To address these limitations, in this work, we introduce JT-DA-8B (JiuTian Data Analyst 8B) with a tool-integrated LLM framework for table reasoning, designed to enhance LLMs’ structural awareness and symbolic reasoning capabilities over table-related tasks through large-scale supervised fine-tuning (SFT) and reinforcement learning (RL). 
As shown in Figure~\ref{fig:overall}, an efficient data synthesization method is proposed that generates large‑scale, high‑quality training examples covering diverse table reasoning tasks. 
Using this synthetic corpus, we then train a purpose‑built large language model that internalizes tool-integrated table reasoning patterns. 
Next, we design a workflow‑based inference architecture that integrates external tools into the reasoning process, allowing the model to invoke code interpreters dynamically and interpret their outputs as part of a multi‑step reasoning chain. 
A visualization tool is also devised to stabilize LLM visualization for tabular data.
Finally, extensive experiments on multiple table reasoning benchmarks are conducted and demonstrate that our approach significantly outperforms existing state‑of‑the‑art models in table reasoning tasks.

Our contributions are as follows:
\begin{itemize}
    % \item A pipeline that programmatically generates millions of high‑quality, diverse training examples is proposed, covering table understanding, table basic operation, table computational operation, data analysis and advanced data analysis.
    \item An architecture for the systematic construction of training data for LLMs in table reasoning is proposed, covering table understanding, table basic operation, table computational operation, data analysis and advanced data analysis, with a pipeline that programmatically generates millions of high‑quality, diverse training samples.
    \item We establish a training paradigm for LLM tool integrated table reasoning capabilities for both SFT and RL. A LLM is trained on our synthetic corpus so that it internalizes tool-integrated table reasoning patterns, markedly reducing table structural sensing misalignments and table reasoning errors.
    \item An effective inference workflow is devised in which tables are preprocessed and the model dynamically calls code interpreters for precise computation, and assembles an interpretable reasoning trace.
    % \item We establish a comprehensive taxonomy of table reasoning tasks, covering the full spectrum from basic structural understanding to advanced analytical reasoning, and provide a formalized representation for the table reasoning workflow.
    \item Extensive experiments on multiple table reasoning tasks demonstrate that our LLM with the framework outperforms existing methods.
    \item The model parameters and workflow of JT-DA are open-sourced, providing practical tools and resources for developers and researchers working on complex table reasoning tasks.
\end{itemize}

\section{Data}
% Xiaofan Liu, Jingjing Yang, Zhiyan Song

\begin{figure*}[htbp]
\centerline{\includegraphics[width=1.0\textwidth]{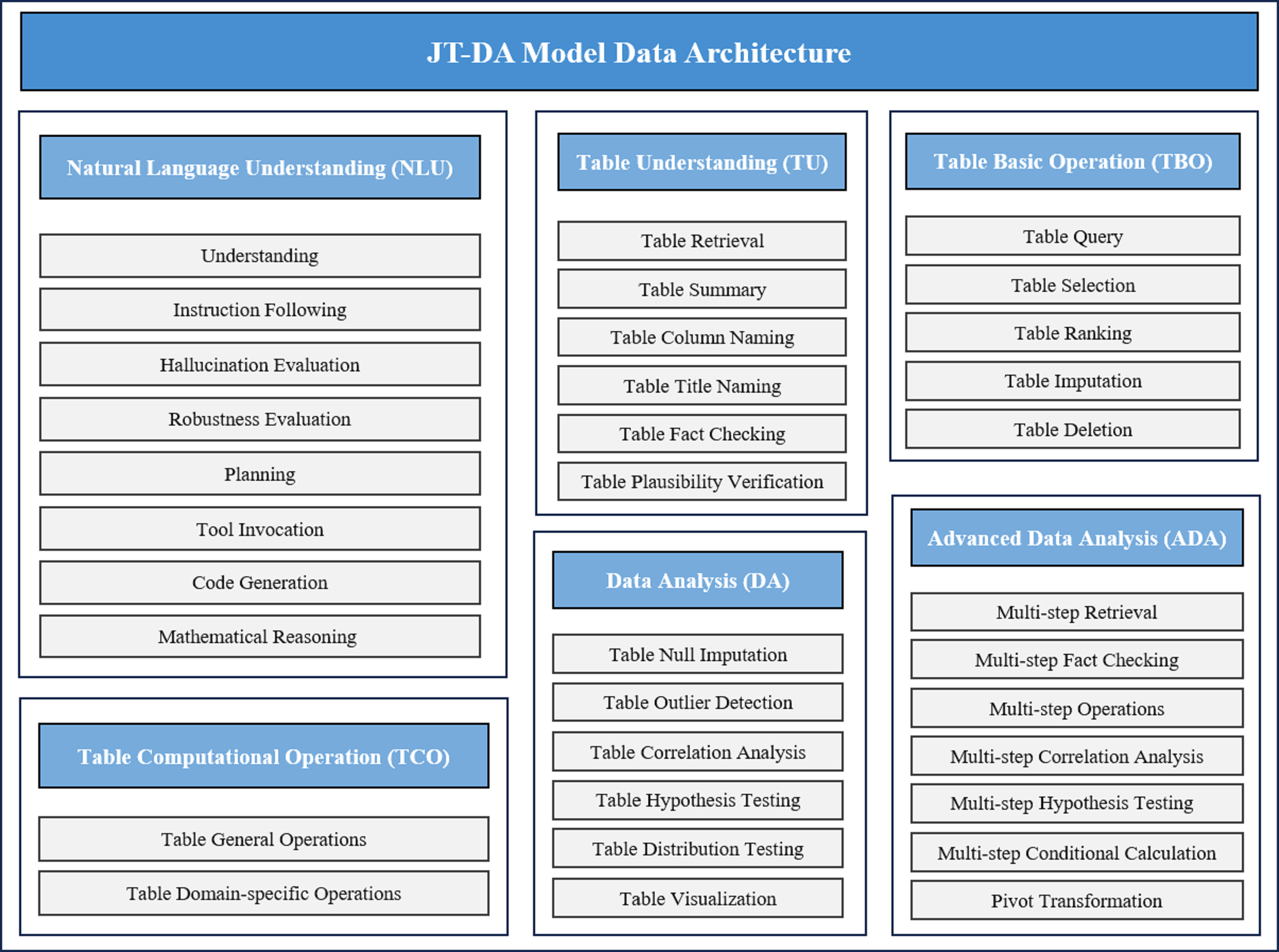}}
\caption{Data architecture overview.}
\label{fig:data_overview}
\end{figure*}

\subsection{Data Overview}

% Currently, the evaluation of large language models (LLMs) in the domain of structured data processing remains constrained by several critical limitations. Prevailing evaluation paradigms continue to rely heavily on single task–oriented datasets. Although recent efforts have introduced multi-task integrated benchmarks, these remain insufficient in both the breadth of data integration and the diversity of task coverage, thereby limiting their capacity to comprehensively assess the ability of LLM generalization in complex structured scenarios. To overcome these shortcomings, this study proposes a novel multi-dimensional hierarchical evaluation framework that spans the full spectrum of capabilities—from fundamental language understanding to advanced intelligent decision-making applications. The framework, as illustrated in Figure \ref{fig:data_overview}, systematically delineates six core capability dimensions and 34 well-defined subtasks, offering a more comprehensive and rigorous basis for assessing LLM performance in structured data contexts.

% The performance of Structured Data Large Language Models heavily depends on the organization and capability coverage of the training data. 
The performance of LLMs on table reasoning tasks is largely determined by the organization and capability coverage of the training data. 
However, existing data constructions suffer from three major limitations. 
%强调任务类型单一
First, task types are highly monolithic. 
Most existing datasets focus solely on table question answering or table retrieval task, which hinders the models' capability of jointly learning heterogeneous tasks such as table understanding, manipulation, computation, and analytical reasoning. 
Second, natural language understanding and table reasoning are typically trained separately on different datasets and in different stages. 
This separation hinders models from simultaneously achieving strong language comprehension and table reasoning. 
% 强调数据集没有系统性的组织形式
Third, there is a significant lack of systematic organization and taxonomy of table tasks and capabilities, along with insufficient diversity in data sources, task types, and capability levels.
This deficiency substantially limits the model’s ability to generalize to complex, real-world table reasoning scenarios that require the integration of multiple analytical skills.
% These limitations collectively lead to the difficulty for current models in forming systematic, composable reasoning and decision-making capabilities for real-world business scenarios.
Overall, these limitations collectively impede LLMs from developing systematic reasoning and decision-making capabilities tailored to real-world business scenarios.

% To address these issues, this paper proposes a Capability-Driven Training Data Construction Framework for Structured Data Large Language Models. This architecture establishes a multi-dimensional, hierarchical training data system, organizing the training data into 6 core capability dimensions and 34 fine-grained sub-tasks, as illustrated in Figure \ref{fig:data_overview}. It covers the complete capability spectrum, from natural language understanding and table understanding and manipulation to statistical analysis and multi-step decision-making, thus providing a structured data foundation for the model's instruction alignment and capability characterization.
To address the aforementioned challenges, this technical report proposes an architecture for constructing training data for LLMs specialized in table reasoning, as depicted in Figure \ref{fig:data_overview}.
The architecture defines a hierarchical taxonomy of table reasoning tasks, systematically organizing them into 6 major capabilities and 34 specific sub-tasks. This taxonomy ensures comprehensive coverage of essential abilities, such as natural language understanding, table understanding, basic and computational table operations, and data analysis. This systematic organization creates the essential data groundwork for training LLMs’ diverse table reasoning capabilities.

% \textbf{Natural Language Understanding (NLU).} This foundational capability dimension systematically evaluates the core competencies of LLMs in natural language processing tasks, encompassing eight critical subtasks, including Understanding, Instruction Following, and Code Generation. Through quantitative analysis of key performance indicators such as parsing precision of linguistic instructions, execution accuracy of operations, generation reliability, hallucination suppression rate, and reasoning consistency, this dimension establishes a multi-dimensional evaluation framework for assessing fundamental language cognition. The evaluation outcomes within this dimension directly reflect the quality of LLM underlying language modeling and provide an essential baseline for subsequent assessments of higher-order capabilities.

% It achieves this through training signals involving intent and key information parsing, alignment of goals and constraints, factual reliability constraints, stability against input perturbations, decomposition of complex goals, external tool selection and parameter construction, mapping from requirements to code, and numerical and formula derivation.
\textbf{Natural Language Understanding (NLU).} This capability covers the core training requirements for the general capabilities of LLMs. It primarily includes key sub-tasks including Understanding, Instruction Following, Code Generation, Hallucination Evaluation, Robustness Evaluation, Mathematical Reasoning, Planning and Tool Invocation. 
This category of data constitutes the foundation of the LLMs' general capabilities, ranging from understanding to controllable generation. 
Specifically, the datasets improve the model's capability of:
\begin{itemize}
    \item Understanding: accurately understanding user intent;
    \item Instruction Following: following instructions while respecting constraints;
    \item Code Generation: generating correct code;
    \item Hallucination Evaluation: maintaining factual reliability;
    \item Robustness Evaluation: remaining robust under input variations or perturbations;
    \item Mathematical Reasoning: generating correct numerical and symbolic reasoning;
    \item Planning: decomposing complex tasks into structured reasoning steps;
    \item Tool Invocation: selecting and configuring external tools appropriately.
\end{itemize}

% \textbf{Table Understanding (TU).} This capability dimension is dedicated to evaluating LLM’s multi-faceted understanding of structured tabular data, encompassing table structure parsing, semantic content comprehension, and validation of table integrity. It comprises six core subtasks—Table Retrieval, Table Summary, Table Column/ Title Naming, Table Fact Checking, and Table Plausibility Verification—designed to assess, respectively, the accuracy of table topology recognition, the completeness of key information extraction, the semantic fidelity of content summarization, the effectiveness of fact-based reasoning, and the integration of domain-specific knowledge. The quantitative assessment of these competencies provides critical empirical evidence for validating the feasibility and robustness of structured question answering (structured QA) systems.

% \textbf{Table Understanding (TU).}  This capability addresses the needs for multi-level semantic and structural understanding of tables. It mainly encompasses Table Retrieval, Table Summary, Table Column Naming, Table Title Naming, Table Fact Checking, and Table Plausibility Verification. These data provide supervision for cell-level evidence localization and fact extraction, global information compression and natural language expression, column-level concept induction and naming, and table-level topic summarization. They serve to enhance the LLMs' capabilities in reading, abstracting, and semantically annotating table content.
\textbf{Table Understanding (TU).}  This capability focuses on developing LLMs' semantic and structural comprehension of tabular data. 
It encompasses a range of tasks, including Table Retrieval, Table Summary, Table Column Naming, Table Title Naming, Table Fact Checking, and Table Plausibility Verification. 
They serve to enhance the LLMs' capabilities in reading, abstracting, and semantically annotating table content.
Specifically, the corresponding datasets provide supervision that improves the model's capability of:
\begin{itemize}
    \item Table Retrieval: cell-level evidence localization and fact extraction;
    \item Table Summary: global information compression and natural language expression for tables;
    \item Table Column Naming: column-level concept induction and naming;
    \item Table Title Naming: table-level topic summarization.
    \item Table Fact Checking: table content verification against external statements;
    \item Table Plausibility Verification: detection of inconsistencies and unlikely patterns within tables.
\end{itemize}

% \textbf{Table Basic Operation (TBO).} This capability dimension focuses on assessing the effectiveness of transforming natural language inputs into structured operational commands. It comprises five fundamental subtasks—Table Query, Table Selection, Table Ranking, Table Imputation, and Table Deletion. Key performance indicators include the accuracy of intent understanding, the precision of field localization, the completeness of condition parsing, and the reliability of data manipulation execution. These metrics collectively evaluate the operational proficiency of LLM in basic data retrieval, multi-condition filtering, dynamic sorting, missing value imputation, and record maintenance. The results obtained in this dimension serve as foundational indicators for determining the processing maturity of LLM in structured data handling and provide an essential capability foundation for supporting more complex computational tasks.

% \textbf{Table Basic Operation (TBO).} This capability focuses on the alignment learning between natural language and structured operations. It covers basic sub-tasks such as Table Query, Table Selection and Table Ranking, Table Imputation, and Table Deletion. Through training objectives such as precise localization and intent-driven querying, condition parsing and set filtering, constrained sorting execution, structured completion and addition, and refined deletion and modification, this data equips the model with executable operational capabilities and result consistency control for interactive table scenarios.
\textbf{Table Basic Operation (TBO).} This capability focuses on the alignment between natural language instructions and structured table operations. 
It covers fundamental sub-tasks including Table Query, Table Selection, Table Ranking, Table Imputation, and Table Deletion. 
TBO supports the LLMs in executing structured operations, maintaining table consistency, and reasoning across interactive table scenarios.
Specifically, the corresponding datasets provide supervision that improves the model’s capability in:
\begin{itemize}
    \item Table Query: cell-level localization and intent-driven retrieval;
    \item Table Selection: conditional filtering and set-based selection;
    \item Table Ranking: constrained sorting and ordering operations;
    \item Table Imputation: structured completion and missing value handling;
    \item Table Deletion: controlled removal and modification of table content.
\end{itemize}
% Through training objectives such as precise localization and intent-driven querying, condition parsing and set filtering, constrained sorting execution, structured completion and addition, and refined deletion and modification, this data equips the model with executable operational capabilities and result consistency control for interactive table scenarios.

% \textbf{Table Computational Operation (TCO).} This advanced capability dimension systematically evaluates LLM’s proficiency in performing complex computations within structured data contexts, encompassing both Table General Operations and Table Domain-specific Operations subtasks. Quantitative metrics—such as the accuracy of mathematical expression compilation, the precision of function mapping, and the effectiveness of domain knowledge integration—are employed to assess the practical performance of LLM in multi-step computational instruction execution, specialized domain calculations, and the construction of automated analytical workflows. The evaluation outcomes in this dimension directly reflect the technological maturity of LLM for applications in vertical industries.

\textbf{Table Computational Operation (TCO).} %This capability addresses operator-level computational needs in structured scenarios,
% This capability is designed to model fine-grained computational operations over structured data,
% primarily consisting of two sub-tasks: Table General Operations and Table Domain-specific Operations. The former centers on single-step calculations for common statistical aggregations on table data, providing supervision for basic numerical analysis. 
% %The latter introduces domain formulas or user-defined formulas to reinforce the model's field alignment and prior knowledge fusion capabilities in calculating specific metrics, thereby supporting the model's transition from mastering general operators to industry-specific computations.
% By incorporating domain-specific or user-defined formulas, this component enhances the model’s alignment with professional fields and integrates prior knowledge. This enables the model to transition from executing general operations to performing industry-specific computations.
TCO aims to develop fine-grained computational operations over structured data.
It comprises two major sub-task families: Table General Operations, which focus on single-step numerical computations and common statistical aggregations, and Table Domain-specific Operations, which incorporate field-dependent or user-defined formulas.
Together, these tasks strengthen the model’s ability to execute precise numerical transformations, align its computation patterns with professional analytical practices, and generalize from generic tabular operations to domain-specific analytical operations.
Specifically, the corresponding datasets improve the model's capability in:
\begin{itemize}
    \item Table General Operations: single-step statistical calculations over tabular values;
    \item Table Domain-specific Operations: domain-aligned formula calculations over tabular values.
\end{itemize}

% \textbf{Data Analysis (DA).} This capability dimension systematically evaluates LLM’s fundamental statistical analysis competencies in structured data contexts, comprising six representative subtasks: Table Null Imputation, Table Outlier Detection, Table Correlation Analysis, Table Hypothesis Testing, Table Distribution Testing, and Table Visualization. Key evaluation criteria include the scientific rigor of missing data handling methods, the sensitivity and specificity of anomaly detection, the statistical power of variable association modeling, the inferential validity of hypothesis testing, the goodness-of-fit in distribution modeling, and the cognitive effectiveness of visual representations. Collectively, these measures provide a comprehensive assessment of the systematic capacity of LLM for statistical insight extraction from structured data. The outcomes of this evaluation signify a critical capability threshold, marking the progression of LLM from basic data processing toward scientific decision support.

\textbf{Data Analysis (DA).} 
% This capability targets foundational statistical analysis over structured data, encompassing core tasks such as Table Null Imputation, Table Outlier Detection, Table Correlation Analysis, Table Hypothesis Testing, Table Distribution Testing, and Table Visualization. Through these tasks, the model is trained to acquire competencies in missing value imputation and data consistency correction, outlier pattern recognition, statistical dependency analysis among variables, significance and difference inference, distributional hypothesis identification, and data-to-visual representation mapping. Together, these tasks establish a coherent supervision framework that supports the model’s learning process from data quality control and anomaly handling to statistical inference and visual result interpretation.
DA focuses on fundamental statistical analysis over structured data, encompassing core tasks including Table Null Imputation, Table Outlier Detection, Table Correlation Analysis, Table Hypothesis Testing, Table Distribution Testing, and Table Visualization. 
Theses tasks establish a comprehensive supervision that supports the model's fundamental ability from data quality assurance and anomaly remediation to robust statistical inference and effective visualization.
Specifically, the datasets provide supervisory signals that improve the model's capability of:
\begin{itemize}
    \item Table Null Imputation: strategies for missing-value completion;
    \item Table Outlier Detection: identification of abnormal patterns and deviations from expected structure;
    \item Table Correlation Analysis: characterization of statistical dependencies among columns;
    \item Table Hypothesis Testing: assessment of inferential validity in tables;
    \item Table Distribution Testing: determination of distributional properties;
    \item Table Visualization: rendering table data into interpretable statistical visualizations.
\end{itemize}

\textbf{Advanced Data Analysis (ADA).}  
%This capability targets composite analysis pipeline design in real-world business contexts. 
% This capability targets multi-step analytical reasoning over structures data in real-world business contexts.
% It covers sub-tasks such as Multi-step Retrieval, Multi-step Fact Checking, Multi-step Operations, Multi-step Correlation Analysis, Multi-step Hypothesis Testing, Multi-step Conditional Calculation, and Pivot Transformation.
% This category emphasizes the consistency of processes in intermediate table construction and result inference, as well as the integration of evidence across different steps. It enables the model to maintain stable analytical trajectories and produce verifiable final outcomes in complex scenarios that involve multi-condition filtering, the creation of derived variables, rule modifications, and multi-dimensional aggregation and reordering.
ADA capability targets multi-step analytical reasoning over structured data, where answers depend on a sequence of intermediate table transformations and evidence integrations.
It covers sub-tasks including Multi-step Retrieval, Multi-step Fact Checking, Multi-step Operations, Multi-step Correlation Analysis, Multi-step Hypothesis Testing, Multi-step Conditional Calculation, and Pivot Transformation.
These sub-tasks can enhance the model's proficiency in coherent data analysis and consistency across intermediate reasoning states so as to support complex table reasoning tasks.
In detail, ADA strengthens the LLMs' capability in:
\begin{itemize}
    \item Multi-step Retrieval: locating required information within tables via multi-hop state tracking;
    \item Multi-step Fact Checking: verifying claims through sequential evidence accumulation;
    \item Multi-step Operations: executing ordered table transformations and calculations;
    \item Multi-step Correlation Analysis: identifying relational patterns across multiple steps;
    \item Multi-step Hypothesis Testing: evaluating hypothesis validity through iterative construction;
    \item Multi-step Conditional Calculation: performing sequential computations where conditions or thresholds may change dynamically across reasoning hops;
    \item Pivot Transformation: reorganizing data layouts while preserving consistency across derived intermediate tables.
\end{itemize}
% This category emphasizes the consistency of processes in intermediate table construction and result inference, as well as the integration of evidence across different steps. It enables the model to maintain stable analytical trajectories and produce verifiable final outcomes in complex scenarios that involve multi-condition filtering, the creation of derived variables, rule modifications, and multi-dimensional aggregation and reordering.

%This category emphasizes process consistency in "intermediate table construction-to-result inference" and cross-step evidence integration capabilities. It enables the model to maintain stable analysis paths and verifiable final outputs in complex scenarios involving multi-condition filtering, derived variables, rule changes, and multi-dimensional aggregation and reordering.

% The proposed evaluation framework adopts a progressive architectural design that not only quantifies the independent performance of each capability dimension using standardized metrics, but also emphasizes the assessment of synergistic effects and compositional gains across capability tiers. This multi-dimensional coupled evaluation paradigm simulates compound requirements encountered in real-world application scenarios, enabling precise appraisal of LLM’s technology implementation value and delivering an accurate capability maturity profile for industrial-scale deployment.

Overall, the capability taxonomy above establishes a comprehensive blueprint of table reasoning abilities from basic operations to multi-step analytical reasoning.
Compared to existing work, the core innovations of this data architecture lie in:
\begin{itemize}
    % \item Capability-Driven Data Design Paradigm: Starting from the table-based application scenarios, we explicitly define the target capabilities the model must possess. We then divide these capabilities into specific sub-tasks and construct corresponding training data around them.
    \item Capability-Driven Data Construction Paradigm: We systematically derive essential table reasoning capabilities from real-world scenarios, decompose them into fine-grained sub-tasks, and subsequently curate a tailored training dataset for LLMs.
    % \item Progressive Capability Construction Path: The task design differentiates various difficulty levels, progressing from simple table retrieval and basic operations to standard data analysis, and finally to multi-step advanced data analysis. This supports the hierarchical and gradual formation of the model's capabilities in structured scenarios.
    \item Progressive Capability Construction: We architect a curriculum of tasks with graduated difficulty, ranging from simple table retrieval and basic manipulation to standard and advanced multi-step data analysis. This design supports the incremental development of the model’s reasoning capabilities over tabular data.
    \item Three Chain-of-Thought (CoT) Data Augmentation Mechanisms: 
    % We introduce differentiated CoT annotation strategies for different capability levels: 
    Differentiated CoT annotation strategies are introduced for tasks at different levels. We employ Textual Chain-of-Thought (TCoT) for the reasoning path in Table Understanding tasks,  Program-of-Thought (PoT) for the inference process in  Data Analysis tasks, and Interleaved Chain-of-Thought (ICoT) for the multi-round iterative reflection process in Advanced Data Analysis tasks. 
    % This strategy enhances the model's interpretability and robustness in complex structured tasks.
    This strategy collectively enhance the interpretability and robustness of LLMs in complex table reasoning tasks.
\end{itemize}

\begin{figure*}[tbp]
\centerline{\includegraphics[width=0.8\textwidth]{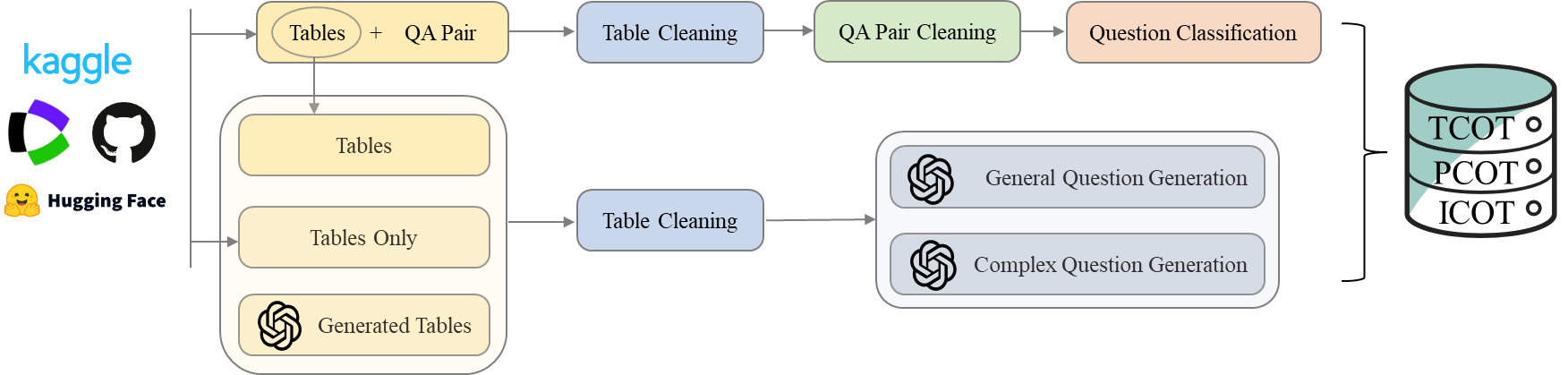}}
\caption{Data construction pipeline.}
\label{fig:data_gen}
\end{figure*}

\textbf{Data Construction Pipeline.} To construct a training dataset with a consistent structure and controllable data type specification, as depicted in Figure \ref{fig:data_gen}, we establish a unified data pipeline for table understanding and reasoning. 
%We acquire real-world tables and QA pairs from multiple sources and introduce synthetic tables to expand coverage.
We collect real-world tables and question–answer pairs from diverse sources, and further introduce synthetically generated tables to broaden task and distributional coverage.
We first perform table standardization cleaning and quality filtering, followed by consistency verification and task type classification for the QA samples. For tables lacking QA, we generate questions of varying complexity. Finally, we perform Chain-of-Thought (CoT) completion to meet the requirements for enhancing table reasoning capabilities.

\subsection{Data Collection}
% Existing public datasets for structured data question answering exhibit notable limitations in both quality and coverage. First, there is a gap between general language understanding and table reasoning. Most models lack joint training across natural language processing and table reasoning, resulting in suboptimal performance on text comprehension and instruction-following tasks. Second, task coverage is limited. Current table QA datasets focus on single-task, hindering multi-task training and cross-task generalization. Third, data sources and structures are homogeneous. Datasets are primarily derived from wiki-style webpages, with limited domain diversity, table size variation, and content richness, which restricts model adaptability across scenarios. To address these limitations, this work proposes a multi-source heterogeneous data construction strategy focusing on three aspects: enhancing language capability, broadening task diversity in table QA, and augmenting non-QA table data.
Existing public datasets for table reasoning exhibit significant limitations in both quality and coverage. 
First, there is a gap between general language understanding and table reasoning. 
Most models lack joint training across natural language processing and table reasoning, resulting in suboptimal performance on table related tasks. 
Second, task coverage is limited. Current table QA datasets primarily focus on single-task, hindering multi-task training and cross-task generalization. 
Third, data sources and structures are homogeneous. 
Existing datasets are predominantly derived from wiki-style webpages with limited domain diversity, table size variation, and content richness, which restricts model adaptability across scenarios. 
To address these limitations, this work constructs a multi-source heterogeneous dataset for table reasoning, focusing on three key aspects: enhancing language capability, broadening task diversity in table QA, and augmenting non-QA table data.

\textbf{Enhancing Natural Language Data.}
% To improve the language understanding ability of structured data models, particularly for tasks involving mathematical reasoning and code generation, we systematically collect and integrate diverse training datasets. On one hand, we select representative datasets covering general language comprehension, reasoning, robustness, instruction following, and factual verification, including MMLU~\cite{wang2024mmlu}, Winogrande~\cite{sakaguchi2021winogrande}, UHGEval~\cite{liang2023uhgeval}, FollowEval~\cite{jing2023followeval}, and AdvGLUE~\cite{wang2021adversarial}. On the other hand, we strengthen datasets related to mathematical and programming capabilities, integrating MATH~\cite{hendrycks2021measuring}, GSM8K~\cite{cobbe2021training}, MathBench~\cite{liu2024mathbench}, BigCodeBench~\cite{zhuo2024bigcodebench}, and DS-1000~\cite{lai2023ds}. All selected datasets undergo rigorous deduplication, format standardization, and multilingual adaptation to ensure high-quality and broadly applicable training data.
To improve the language understanding ability of our model, particularly for tasks involving mathematical reasoning and code generation, we systematically collect and integrate diverse training datasets. 
On one hand, we select representative datasets covering general language comprehension, reasoning, robustness, instruction following, and factual verification, including MMLU~\cite{wang2024mmlu}, Winogrande~\cite{sakaguchi2021winogrande}, UHGEval~\cite{liang2023uhgeval}, FollowEval~\cite{jing2023followeval}, and AdvGLUE~\cite{wang2021adversarial}. 
On the other hand, to strengthen mathematical and programming capabilities, we incorporate datasets such as MATH~\cite{hendrycks2021measuring}, GSM8K~\cite{cobbe2021training}, MathBench~\cite{liu2024mathbench}, BigCodeBench~\cite{zhuo2024bigcodebench}, and DS-1000~\cite{lai2023ds}. 
All selected datasets undergo rigorous deduplication, format standardization, and multilingual adaptation to ensure high-quality and broadly applicable training data.

\textbf{Broadening Task Diversity in Table QA.}
% To enhance model generalization on downstream tasks, we extend the data collection scope from standard table QA to diverse tasks such as table summary, cell-level QA, table fact checking, table structure recognition, column annotation, and table-based data analysis. Based on dataset representativeness, timeliness, format diversity, and redundancy, we conduct strict selection and identify 29 representative datasets, including AIT-QA~\cite{aitqa} , ToTTo~\cite{totto}, HybridQA~\cite{hybridqa}, and TableBench~\cite{wu2025tablebench}.
To enhance model generalization on downstream tasks, we extend the data collection scope beyond standard table QA to incorporate diverse tasks such as table summary, cell-level QA, table fact checking, table structure recognition, column annotation, and table-based data analysis. 
According to dataset representativeness, timeliness, format diversity, and redundancy, we conduct a rigorous selection process and identify 29 representative datasets, such as AIT-QA~\cite{aitqa}, ToTTo~\cite{totto}, HybridQA~\cite{hybridqa}, and TableBench~\cite{wu2025tablebench}.

\textbf{Augmenting Non-QA Table Data.}
% To increase the diversity and coverage of structured data, we systematically collect pure table data without question-answer pairs. According to structural characteristics, we categorize tables into two types: CSV files with regular structure and Excel files with multi-level headers or merged cells. Data collection leverages domain-specific keywords on platforms such as Web of Science and GitHub, followed by cleaning and normalization. To ensure domain coverage and representativeness, we include tables from multiple domains: economics (e.g. income statements), business operations (e.g. inventory management), telecommunications (e.g. 5G base station deployment, tariff comparisons), and education (e.g. student performance). All collected tables strictly adhere to the following quality and format standards:
To enhance the diversity and coverage of tabular data, we systematically collect pure table data without associated question-answer pairs. 
According to structural characteristics, tables are categorized into two types: CSV files with regular structured content and Excel files with semi-structured content. 
Data collection leverages domain-specific keywords on platforms such as Web of Science and GitHub, followed by comprehensive cleaning and normalization. 
To ensure broad domain coverage and representativeness, the dataset includes tables from multiple domains, including economics (e.g. income statements), business operations (e.g. inventory management), and education (e.g. student performance). All collected tables strictly comply with the following quality and format standards:
% \begin{itemize}
%     \item Tables have non-empty first row and first column;
%     \item Content consists of parseable numbers or text, excluding images or garbled data; 
%     \item Individual file size does not exceed 100 MB;
%     \item Table body is not entirely empty beyond headers; 
%     \item In single-row header tables, every column has a clear name. 
% \end{itemize}
\begin{itemize}
    \item Tables have non-empty first row and first column;
    \item Table content consists solely of parseable numbers or text, excluding images or corrupted data; 
    \item Individual file does not exceed 100 MB in size;
    \item Table bodies contain at least one non-empty row beyond the headers; 
    \item In single-row header tables, every column is assigned a well-defined and unambiguous header name. 
\end{itemize}
% This study constructs a dataset comprising 3 million tables, achieving high standards in domain coverage, format consistency, and data quality.
This work constructs a dataset comprising approximately 3 million tables, ensuring extensive domain coverage, format consistency, and high data quality.

\subsection{Data Cleaning}
\begin{figure*}[tbp]
\centerline{\includegraphics[width=0.8\textwidth]{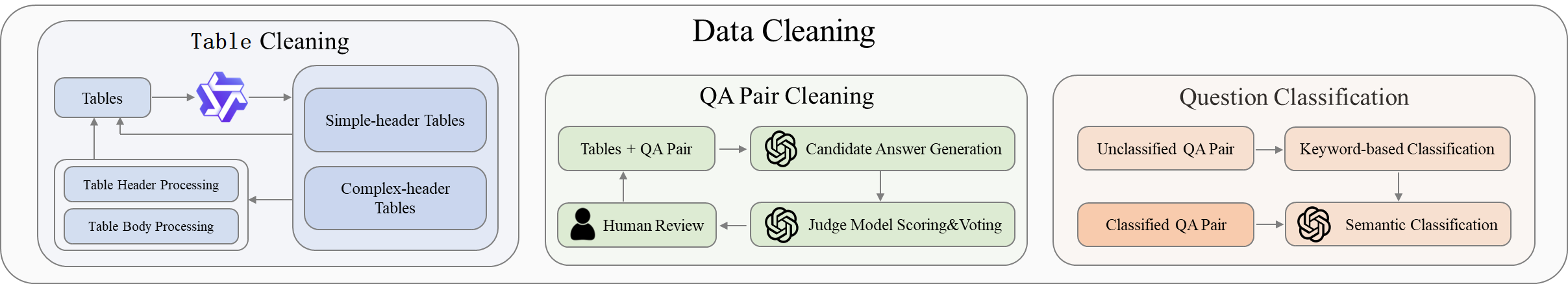}}
\caption{Data cleaning pipeline.}
\label{fig:data_clean}
\end{figure*}

% To ensure the accuracy and usability of the dataset, this study implements a multi-level data cleaning pipeline encompassing table cleaning, QA pair cleaning, and task question classification. The pipeline integrates automated rules, large language model reasoning, and human review, aiming to maximize data quality while minimizing noise.
To ensure dataset accuracy and usability, a multi-level data cleaning pipeline is devised in this work, encompassing table cleaning, QA pair cleaning, and task question classification. 
As depicted in Figure \ref{fig:data_clean}, the pipeline combines automated rules, LLM reasoning, and human review, systematically maximizing data quality while minimizing noise.

\textbf{Table Cleaning.}
% Tables consist of two parts: headers and body content, which differ significantly in structure and cleaning requirements and therefore require separate processing. Headers provide a high-level summary of the table content and are categorized into simple and complex headers. Simple-header tables contain a single row without merged cells, whereas complex-header tables include multiple rows or merged cells. This study constructs a table classification model based on Qwen3-32B model, achieving 94.21\% accuracy in identifying complex headers. For complex headers, the processing workflow involves: (i) splitting merged cells; and (ii) merging multi-row headers into a single row using underscores to enable uniform parsing.
Tables consist of two parts: headers and body content, which differ substantially in structure and cleaning requirements, necessitating separate processing strategies. 
First, header processing begins with the identification of complex-header tables. 
Headers usually provide a high-level summary of the table content and can be categorized into simple and complex headers. 
Simple headers consist of a single row without merged cells, whereas complex headers may span multiple rows or include merged cells. 
In order to classify the tables, we construct a table classification model based on Qwen3-32B, achieving 94.21\% accuracy in identifying complex headers. 
Afterwards, for complex-header tables, the processing workflow involves: (i) splitting merged cells; and (ii) merging multi-row headers into a single row using underscores to enable uniform parsing.

% The table body may contain high proportions of missing values, mixed data types, or inconsistent formats. The cleaning procedure includes: (i) removing rows containing only annotations; (ii) eliminating superscripts and subscripts within cells; (iii) converting URLs, Excel formulas, and similar content to plain text; and (iv) discarding tables with more than 70\% missing values to ensure usability and completeness.

Second, the table body may contain high proportions of missing values, mixed data types, or inconsistent formats which requires multiple cleaning processes. 
The major cleaning procedures include: (i) removing rows containing only annotations; (ii) eliminating superscripts and subscripts within cells; (iii) converting URLs, Excel formulas, and similar content to plain text; and (iv) discarding tables with more than 70\% missing values to ensure usability and completeness.

In summary, the table cleaning process ensures that the tables are consistently formatted and reliable. 
This rigorous preprocessing not only improves data quality and parsing efficiency but also provides a solid foundation for subsequent data synthesization and LLM training.

\textbf{QA Pair Cleaning.}
% To address errors and inconsistencies in publicly available table QA datasets, this study implements a multi-stage quality control framework:
To address errors and inconsistencies in publicly available table QA datasets, a multi-stage quality control framework is implemented in this work:

% \begin{itemize}
%     \item Candidate answer generation: Three large language models generate candidate answers for non-rule-based QA pairs.
%     \item Content consists of parseable numbers or text, excluding images or garbled data; 
%     \item Judge model scoring and voting: A judge model evaluates the consistency between the three candidate answers and the ground truth, assigning quality scores based on answer agreement. Scores of 10, 9, and 8 indicate high-score QA pairs, whereas scores of 2 and 0 indicate low-score pairs.
%     \item Human review: Non-perfect QA pairs undergo sampled human inspection, with domain experts verifying and correcting answers as needed.
% \end{itemize}

\begin{itemize}
    \item Candidate answer generation: 3 different kinds of LLMs are used to generate candidate answers for non-rule-based QA pairs.
    \item Content filtering: Only parseable numbers or text are retained, excluding images or corrupted data. 
    \item Judge model scoring and voting: A judge model is used to evaluate the consistency among the 3 candidate answers and the ground truth, assigning quality scores based on answer agreement. 
    Higher scores indicate higher data quality.
    \item Human review: Imperfect QA pairs are subjected to sampled human inspection, with domain experts verifying and correcting answers as needed.
\end{itemize}
% Finally, QA data are divided into high-quality data (rule-generated, human-verified, or high-scoring QA pairs) and low-quality data (unreviewed or low-scoring QA pairs ).

Finally, the QA data are categorized into high-quality data (rule-generated, human-verified, or high-scoring QA pairs) and low-quality data (unreviewed or low-scoring QA pairs). 
Low-quality data are filtered out in this stage.

\textbf{Question Classification.}
% To improve the applicability of the data for downstream tasks, QA pairs are reclassified according to six defined capabilities and 34 subtask types:
To improve the utility of the data for downstream tasks, QA pairs are reclassified according to 6 well-defined capabilities encompassing 34 subtask types. We adopt 2 different kinds of classification methods:

% Keyword-based classification: Regular expressions with task-specific keywords identify 21 tasks with clear features, such as Table Summary, Table Column/ Title Naming, Table Fact Checking, Table Query and Table Ranking.

% Semantic classification: Large language models leverage semantic understanding and task boundary definitions to classify 13 tasks that are difficult to distinguish using keywords alone, such as Table Plausibility Verification, Table General Operations, Table Domain-specific Operations, Table Outlier Detection and Multi-step Operations.

\begin{itemize}
    \item Keyword-based classification: Regular expressions with task-specific keywords are used to identify 21 tasks with distinctive features, including Table Summary, Table Column/Title Naming, Table Fact Checking, Table Query and Table Ranking. % TODO
    \item Semantic classification: For the remaining 13 tasks that are difficult to distinguish using keywords alone, LLMs are employed to leverage semantic understanding and task boundary definitions to classify the remaining tasks. This approach covers tasks including Table Plausibility Verification, Table General Operations, Table Domain-specific Operations, Table Outlier Detection and Multi-step Operations. % TODO
\end{itemize}

\subsection{Data Synthesization}
% Analysis of collected Table QA datasets reveals that they primarily focus on table summary and understanding, which remain insufficient to cover all sub-tasks required for training. To address this gap, this study designs a multi-level data generation and augmentation strategy to systematically enhance both task diversity and coverage.

Analysis of existing Table QA datasets indicates that the datasets predominantly target table summarization and understanding, leaving many sub-tasks underrepresented for comprehensive table reasoning. 
To bridge this gap, a multi-level table QA generation and augmentation strategy is proposed in this work to systematically enhance both task diversity and coverage.

% For ten well-defined sub-tasks with relatively consistent question formats, the data generation adopts a rule-based approach, such as Table Retrieval, Table Query, Table Selection, Table Ranking, Table Imputation, Table Deletion, Table Null Imputation, Table Correlation Analysis, Table Hypothesis Testing and Table Distribution Testing. The process consists of: (1) selecting eligible tables containing numeric columns with computable conditions; (2) designing templated rules based on common data manipulation logic and predefined field combinations to automatically generate QA pairs, covering task types such as statistical computation, information retrieval, single- and multi-condition filtering, deletion, and sorting.

For 10 well-defined sub-tasks with relatively consistent question formats—including Table Retrieval, Table Query, Table Selection, Table Ranking, Table Imputation, Table Deletion, Table Null Imputation, Table Correlation Analysis, Table Hypothesis Testing and Table Distribution Testing—we adopt a rule-based data generation approach. 
The process consists of: (1) selecting eligible tables containing numeric columns with computable conditions; (2) designing templated rules based on common data manipulation logic and predefined field combinations to automatically generate QA pairs.
This procedure covers task types such as statistical computation, information retrieval, single-condition filtering, multi-condition filtering, deletion, and sorting.

% For four clearly defined sub-tasks with relatively consistent question formats, the data generation process applies a rule-based method, such as Table General Operations, Table Domain-specific Operations, Table Plausibility Verification and Table Summary.Two generation modes are used: (1) generating QA pairs directly from existing tables, and (2) generating both tables and corresponding QA pairs in the absence of pre-existing data. The process includes: (1) designing tailored prompts for each sub-task, and (2) incorporating two independent discriminator models to evaluate generated samples on accuracy, relevance, and coverage, retaining only those rated as perfect by both discriminators. This dual-verification mechanism ensures data accuracy and usability while maintaining diversity, ultimately producing a high-quality, standardized dataset.

In addition, for 4 clearly defined sub-tasks with relatively diverse question formats (including Table General Operations, Table Domain-specific Operations, Table Plausibility Verification, and Table Summary), we employ a complementary LLM-based generation strategy, including two generation modes: (1) generating QA pairs directly from existing tables, and (2) generating both tables and corresponding QA pairs in the absence of pre-existing tables. 
The generation process includes: (1) designing tailored prompts for each sub-task, and (2) applying two independent discriminator models to evaluate generated samples on accuracy, relevance and coverage, retaining only those rated perfect by both. This dual-verification mechanism ensures data quality and usability while maintaining diversity, ultimately producing a high-quality, standardized dataset.

\subsection{Process-Level Data Augmentation}
% We perform Chain-of-Thought (CoT) completion to address a prevalent structural gap in existing training data for table understanding and numerical reasoning: a sufficiency of "question-answer pairs" but a scarcity of "process supervision." When training relies solely on final answers, models tend to depend on spurious correlations or template-based mapping. Consequently, they fail to stably learn general strategies such as multi-step calculation, conditional filtering, and cross-column alignment, leading to reasoning leaps and unreliable generation on complex examples. Therefore, we complete the CoT for the training data. The goal is to enhance data coverage with higher-density, transferable process signals, providing the model with dual supervision: "path-level learning" and "result-level alignment."

We enrich the training corpus with Chain-of-Thought (CoT) annotations to address a common structural gap in our train datasets: while question–answer pairs are abundant, process-level supervision is scarce.
Training solely on final answers encourages models to memorize patterns or rely on spurious correlations, which undermines their ability to generalize to novel examples.
Consequently, they struggle to acquire stable strategies for table reasoning tasks, resulting in reasoning leaps and unreliable outputs on unseen queries. 
To mitigate this issue, we supplement the data with explicit CoT reasoning traces, thereby increasing process-signal density and transferability.
This enhancement provides the model with dual supervision: path-level learning and result-level alignment, ultimately improving its robustness on structured reasoning tasks.

% Based on this objective, we adopt two complementary routes, TCoT (Textual Chain-of-Thought) and PoT (Program-of-Thought), to augment the training data.

Motivated by this objective, we leverage two complementary strategies: Textual Chain-of-Thought (TCoT) and Program-of-Thought (PoT), to augment the training data.

% TCoT serves as a natural language process representation of the tree-like thinking paradigm. It emphasizes the organization of multi-branch candidate ideas, the localization of key evidence, and the gradual convergence of intermediate conclusions. We utilize DeepSeek and Qwen-72B to generate high-quality solutions for target questions. We then perform structural induction and compression on their reasoning processes to distill TCoT text suitable for direct training. This category of completion data primarily provides "verbalized, multi-path, and interpretable" process supervision, enhancing the model's global reasoning organization capabilities and its ability to decompose complex problems.

% PoT serves as the process representation for the programmatic reasoning paradigm. It explicitly transcribes solution steps into executable code, thereby ensuring low-ambiguity and verifiable calculation trajectories. We employ QWQ-32B to generate solution code for questions and execute it within a sandbox to obtain deterministic outputs. These "code-execution result" pairs are then used as training augmentation samples. This category of completion data provides stronger verifiable process supervision for numerical calculations and table operations, improving the model's learning quality regarding computational correctness and multi-step derivation stability.

\begin{itemize}
    \item TCoT serves as a textual chain-of-thought representation for table reasoning. It captures natural-language reasoning patterns by highlighting table-relevant evidence and consolidating intermediate conclusions. 
    To construct this data, we employ both DeepSeek-R1 and Qwen-72B to generate high-quality solutions for table-focused queries. 
    We further reorganize and condense their raw reasoning traces into concise and structurally coherent TCoT rationales suitable for direct training.
    This category of data primarily provides verbalized, multi-path, and interpretable process supervision, enhancing the model's global reasoning organization and its ability to decompose and solve complex table reasoning tasks.
    \item PoT serves as the process representation for the programmatic reasoning paradigm. 
    It explicitly translates solution steps into executable code, thereby yielding low-ambiguity and verifiable computational trajectories. 
    In our pipeline, QWQ-32B is used to generate solution programs, which are executed in a sandboxed environment to obtain deterministic outputs. 
    The resulting question-code pairs are incorporated as augmentation samples during training. 
    This category of data provides stronger verifiable process supervision for table computational operation and data analysis scenarios, substantially enhancing the model’s coding correctness and data analysis reliability.
\end{itemize}

% In summary, we use TCoT to complete training signals for "thought organization and multi-branch reasoning," and PoT to complete signals for "executable and verifiable computation." These two strategies synergistically expand the process coverage and difficulty hierarchy of the training data, systematically improving the model's reasoning robustness and trustworthy output capabilities in complex structured tasks from a data-centric perspective.

In summary, TCoT is employed in our work to enrich thought organization and logical reasoning of LLMs in tabular tasks, while PoT is used to provide precise, executable and verifiable computation traces. 
These two strategies synergistically expand the process coverage and difficulty hierarchy of the training data, thereby systematically enhancing the model’s reasoning robustness and reliability in complex tabular tasks from a data-centric perspective.

\section{Complex Table Dataset Synthesization}
\subsection{Overview}

\begin{figure*}[tbp]
\centerline{\includegraphics[width=0.8\textwidth]{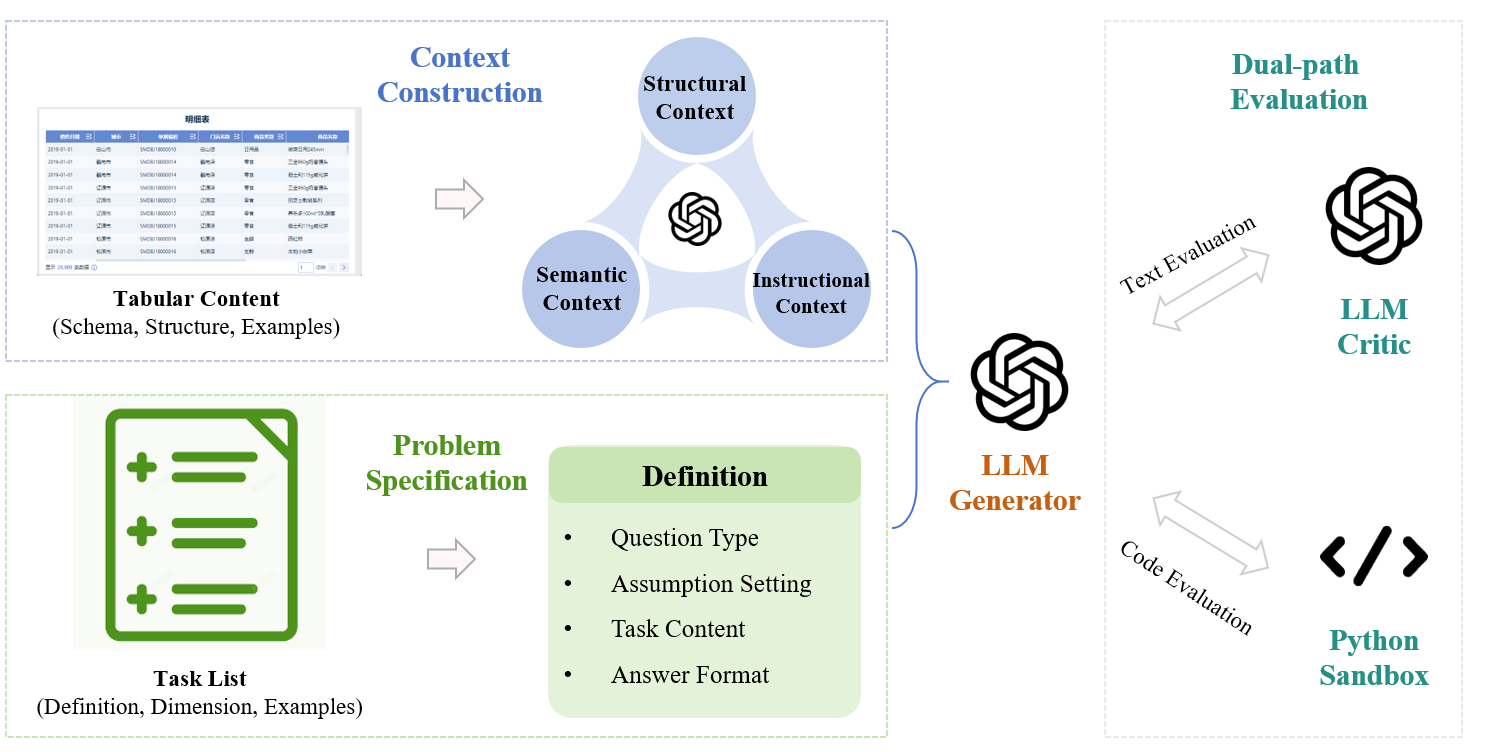}}
\caption{Complex table dataset synthesization.}
\label{fig:ComplexQA}
\end{figure*}

% Real-world data analysis rarely involves simple single-step lookups; instead, it typically requires multi-step reasoning over complex structured tables. 
% However, most existing datasets only cover small-scale tables and shallow question types, which significantly limits the ability of current models to acquire and demonstrate advanced analytical skills. 
% To bridge this gap, we construct a dataset that closely mimics realistic analytical workflows, with the goal of enhancing models’ capability to perform step-by-step reasoning within the CoT paradigm.

Real-world data analysis typically requires multi-hop reasoning over complex structured tables, where analysts need to iteratively integrate information, perform intermediate computations, and verify assumptions. 
However, most existing datasets predominantly focus on small-scale tables and single-hop or shallow queries, offering limited coverage of the reasoning patterns observed in practical analytical workflows.
The mismatch of train datasets and realistic data analysis demands restricts the trained models from multi-hop step-dependent reasoning skills for table reasoning.
To bridge this gap, we introduce a new dataset that closely mimics realistic multi-step analytical processes within the CoT paradigm.
Our dataset features substantially realistic complex tables, deeper reasoning chains, and diverse task types that span retrival, fact checking, operations, correlation analysis, conditional calculation and pivot transformation.
By mirroring the complexity of real-world analytical tasks, our dataset aims to facilitate the development of models capable of performing step-by-step reasoning required for end-to-end data analysis.

% As depicted in Figure \ref{fig:ComplexQA}, our approach follows a structured generation pipeline that unifies task construction, reasoning synthesis, and quality assurance. Specifically, we collect large and semantically rich tables from diverse domains, define analytical tasks along multiple dimensions, and employ powerful large language models to generate multi-step reasoning chains grounded in these tables. To ensure both semantic soundness and executable feasibility, we further adopt a dual-path evaluation mechanism that combines textual critique by LLMs with executable verification in a Python sandbox. Finally, high-quality answers are produced through an iterative sandbox-based reasoning framework and a teacher–student CoT distillation process, which together guarantee correctness, interpretability, and diversity of reasoning traces.

As depicted in Figure \ref{fig:ComplexQA}, our data construction framework follows a structured generation pipeline that unifies task construction, reasoning synthesis, and quality control. 
Specifically, we collect large and semantically rich tables from diverse domains, define analytical tasks along multiple dimensions, and employ powerful LLMs to generate multi-step reasoning chains grounded in these tables. 
To ensure both semantic validity and executable correctness, we further adopt a dual-path verification mechanism that combines LLM-based textual critique with programmatic evaluation in a Python sandbox. 
Finally, high-quality instances are obtained through an iterative sandbox-based reasoning workflow and a teacher–student CoT distillation process, which together guarantee correctness, interpretability, and diversity of the resulting reasoning traces.

% In summary, this pipeline enables the construction of a dataset that is complex in content, rigorous in reasoning, and reliable in execution, providing a solid foundation for training and evaluating models on realistic multi-step tabular analysis tasks.
In summary, this pipeline enables the construction of a dataset that is rich in content, rigorous in reasoning, and reliable in execution.
The resulting traces provide a solid foundation for training and evaluating models on realistic multi-step tabular analysis tasks.

\subsection{Data Collection}

To support the construction of realistic analytical tasks, we curated a large set of tables that reflect the structural and semantic complexity observed in practice. In total, we collected over 1,000 tables spanning domains such as telecommunications, finance, and forecasting, obtained through multiple acquisition channels. Compared with the small and homogeneous tables typically used in existing benchmarks, these resources are significantly richer in scale and diversity.

A table is regarded as \textit{complex} if it satisfies the following criteria:
\begin{itemize}
    \item Column names convey clear physical or semantic meaning.
    \item The table contains more than 100 rows and more than 10 columns.
    \item At least two columns provide filterable textual data.
    \item At least four columns include numerical values with non-zero variance, suitable for computation.
    \item The overall proportion of missing values does not exceed 80\%.
    \item The table avoids noise such as garbled characters or excessively long strings.
\end{itemize}

All collected tables were manually reviewed by trained annotators to ensure compliance with these standards, thereby guaranteeing their reliability and utility for constructing challenging multi-step data analysis tasks.

\subsection{Question Synthesization}

Given an input table $T \in \mathbb{R}^{m \times n}$, we first construct a contextual representation 
$H_T = \{ H_{\text{structure}}, \allowbreak H_{\text{semantics}}, \allowbreak H_{\text{indicators}} \}$, 
which serves as the foundation for controlled question generation. 
Here, $H_{\text{structure}}$ encodes structural features such as column names, data types, and hierarchies; 
$H_{\text{semantics}}$ captures natural language interpretations and inferred inter-column relationships; 
and $H_{\text{indicators}}$ denotes a set of derived metrics or domain-relevant variables 
(e.g., mean, ratio, growth rate) inferred from the table content. 
This contextualization step enables the model to capture not only the table’s format but also its analytical potential.  

On top of this representation, we define a task specification through an instruction tuple 
$I = (q_{\text{type}}, \allowbreak s_{\text{source}}, \allowbreak o_{\text{task}}, \allowbreak y_{\text{format}})$, 
where $q_{\text{type}}$ denotes the question intent (e.g., \textit{Verification}, \textit{Application}), 
$s_{\text{source}}$ specifies the grounding type (\textit{Raw} or \textit{Hypothetical}), 
$o_{\text{task}}$ indicates the analytical objective (e.g., \textit{Retrieval}, \textit{Operations}, \textit{Correlation Analysis}, \textit{Hypothesis Testing}), 
and $y_{\text{format}}$ constrains the answer format (e.g., \textit{SingleValue}, \textit{ListFields}, \textit{Yes/No}). 
Only semantically compatible combinations are preserved. 
The final prompt $P = f(I, H_T)$ is then constructed by integrating this instruction with the contextual representation.  

For question generation, we employ an advanced LLM $\mathcal{M}$ (e.g., \textsc{QwQ}, \textsc{DeepSeek}) 
with strong multi-step reasoning capabilities over structured data. 
Prompted with $P$, the model first proposes a preliminary short-form question $Q^{(0)}$, 
and then self-reflectively develops a reasoning chain 
$S = [s_1, s_2, \dots, s_k]$, 
where each step $s_i$ corresponds to a concrete operation such as filtering, grouping, or aggregation. 
To ensure sufficient reasoning depth, we impose a minimum-step requirement:
\begin{equation}
|S| \geq s_{\min}.
\end{equation}
If the chain is too shallow, the model is required to revise and elaborate the question and reasoning logic.  

Finally, we apply a dual-path evaluation to guarantee both semantic soundness and executable feasibility. 
First, a textual critique function (also based on LLM $\mathcal{M}$) evaluates the clarity, logical consistency, 
and contextual alignment of the tuple $(Q, S, H_T)$. 
Second, the reasoning steps $S$ are compiled into Python code and executed in a sandbox environment on the original table $T$. 
Only when both semantic validation and code execution succeed is the generated question admitted into the final set:
\begin{equation}
\text{Verify}(Q_i, S_i, T) = \text{True} 
\quad \Rightarrow \quad Q_i \in \mathcal{Q}.
\end{equation}

This process ensures that each retained question is grounded in the table’s semantics and structure, 
exhibits sufficient analytical depth, and can be reliably executed over real data.

\subsection{Interleaved Chain Distillation}
To address the complexities inherent in multi-step data analysis, we have introduced the Interleaved Chain of Thought (ICoT) framework. ICoT is an advanced reasoning framework designed to elevate the reasoning capabilities of LLMs from the traditional linear thinking pattern to a dynamic, human-like analytical process. When humans tackle complex problems, their thought process is not unidirectional; rather, it continuously cycles through reasoning (Reasoning), acting (Action), and observing (Observation). 

This cyclical mechanism is crucial for addressing the issue of state drift. State drift refers to the phenomenon where, in multi-turn interactive tasks, a model, lacking memory of intermediate logic and assumptions, performs repetitive and inefficient operations or deviates from its initial goal. ICoT resolves this by embedding a continuous thinking-action loop with an external execution environment, ensuring that the model’s planning, intent and conclusions maintain consistency and coherence throughout the entire reasoning process. This approach enables the model to not only store knowledge (facts) but also maintain dynamic logic for real-time adjustment and iterative decision-making.

Formally, the ICoT training set consists of interleaved chains of thought generated by the model during task resolution. We define the reasoning process as the evolution of a series of reasoning states $s_t$, where $t \in \{1, 2, \dots, T\}$ represents the reasoning steps. Each state transition depends not only on the previous state $s_{t-1}$, but also on the result of the previous tool action $r_{t-1}$ and the current invoked $\text{tool}_t$. The generation of reasoning states can be described by the function $f_{\text{think}}$:

\[
s_t = f_{\text{think}}(s_{t-1}, r_{t-1}, \text{tool}_t) \quad \text{for } t = 1, 2, \dots, T
\]

where $f_{\text{think}}$ represents the model's reasoning mechanism. To enhance the model's robustness and adaptability, we designed two complementary tool interaction modes to generate the result $r_t$:

\begin{itemize}
    \item \textbf{Tool-Call-based Data (Dialogue Mode)}: The result $r_t$ is returned in the form of a structured tool role, clearly distinguishing the boundaries between reasoning and tool output, similar to multi-turn dialogues:
    \[
    r_t = \text{ToolCall}(s_{t-1}) \quad \text{where } r_t \in \text{tool role}
    \]
    \item \textbf{ReAct-based Data (Continuous Chain Mode)}: The result $r_t$ is directly concatenated with the reasoning sequence, forming a continuous flow of thinking and actions. This reinforces the model’s ability to immediately integrate observations and continue reasoning:
    \[
    r_t = \text{ReAct}(s_{t-1}, r_{t-1})
    \]
\end{itemize}

Ultimately, the complete distilled dataset $D_T$ consists of all reasoning states and their corresponding results from $T$ rounds of interaction:

\[
D_T = (\{s_1, r_1\}, \{s_2, r_2\}, \dots, \{s_T, r_T\})
\]

By fine-tuning on $D_T$, the model acquires Progressive Learning capabilities, enabling it to build deeper understanding based on prior reasoning in each round. The implementation of ICoT ensures that the model not only stores ``facts'' (similar to traditional memory) but also maintains dynamic ``logic,'' ensuring that in complex tasks that require real-time adjustments and iterative decision-making, the agent can retain coherence and accuracy. This represents a significant breakthrough over the traditional Chain-of-Thought-based reasoning paradigm, marking a pivotal step towards developing advanced agent systems with dynamic, robust, and self-correcting capabilities.

\subsection{Answer Synthesization}

To generate high-quality answers with sufficient analytical depth, we design a staged mechanism that combines 
an executable sandbox-based iterative reasoning framework with a teacher–student ICoT knowledge distillation process. 
The final dataset thus contains the original question, a verified reference answer, and the optimal ICoT reasoning trace 
from the student model after filtering.  

The sandbox-based iterative reasoning framework integrates a Python sandbox execution environment with the reflective 
capabilities of LLMs, enabling multi-turn, interactive answer generation. 
This design substantially improves both the accuracy and robustness of solutions for complex analytical tasks. 
The framework proceeds in three steps:
\begin{itemize}
    \item \textbf{Table perception initialization:} The system prompt guides the LLM to act as a data analysis assistant, 
    extracting core metadata from the target table, including headers, data types, and sample records.
    \item \textbf{Code generation and execution:} The user query is passed to the LLM, which generates Python code to 
    solve the problem. The code is executed within an isolated sandbox environment.
    \item \textbf{Result verification and refinement:} Execution outputs (or error messages) are returned to the LLM. 
    If the result is valid, the answer is finalized; otherwise, the model diagnoses errors, regenerates the code, 
    and iteratively repeats the process until a reliable solution is obtained.
\end{itemize}

To further enhance reasoning quality and enable capability transfer, we adopt a teacher–student CoT distillation procedure:
\begin{itemize}
    \item \textbf{Teacher model generation:} A high-performing teacher model independently produces five candidate answers for each question.
    \item \textbf{Reference answer selection:} A voting mechanism, supplemented by human verification when necessary, 
    identifies the most reliable candidate as the reference answer.
    \item \textbf{Student model distillation:} The student model generates five ICoT responses 
    under the guidance of the reference reasoning chain.
    \item \textbf{Optimal ICoT filtering:} The generated ICoTs are evaluated against the reference answer, 
    and the one with the highest consistency and clearest reasoning is retained for training.
\end{itemize}

Through this two-stage design, the dataset ensures that answers are not only correct and executable 
but also supported by clear and interpretable reasoning traces.

\section{Supervised Fine-tuning}
% Ce Chi, Zhendong Wang

\subsection{Model Architecture}
JT-DA-8B is built based on JT-Coder-8B model \cite{hao2025jt}, a decoder-only LLM trained from scratch, comprising approximated 8 billion paramters. The model architecture follows the main-stream Transformer decoder design, including RMS layer normalization \cite{zhang2019root}, RoPE \cite{su2024roformer}, and GQA mechanism \cite{ainslie2023gqa}.
Specifically, as shown in Table~\ref{tab:model_config}. JT-DA-8B consists of 32 layers with a hidden size of 4096 and a FFN hidden size of 13312. It employs 32 attention heads with 8 key-value heads for efficient computation and memory reduction based on GQA mechanism. The model is trained with a maximum sequence length of 8192 tokens. 

\begin{table}[h]
\centering
\renewcommand{\arraystretch}{1.2} % 增加行高
\caption{JT-DA-8B model configuration.}
\resizebox{\textwidth}{!}{
\begin{tabular}{
    >{\centering\arraybackslash}p{2.0cm}  % Hidden Size
    >{\centering\arraybackslash}p{3.0cm}  % FFN Hidden Size
    >{\centering\arraybackslash}p{2.0cm}  % Layer Number
    >{\centering\arraybackslash}p{1.8cm}  % Heads
    >{\centering\arraybackslash}p{2.0cm}  % KV Heads
    >{\centering\arraybackslash}p{3.0cm}  % Sequence Length
    >{\centering\arraybackslash}p{3.0cm}  % Vocabulary Size
    >{\centering\arraybackslash}p{3.0cm}  % Parameters Number
}
\toprule
\textbf{Hidden Size} & \textbf{FFN Hidden Size} & \textbf{Layer Number} & \textbf{Heads} & \textbf{KV Heads} & \textbf{Sequence Length} & \textbf{Vocabulary Size} & \textbf{Parameters Number} \\
\midrule
4096 & 13312 & 32 & 32 & 8 & 8192 & 152064 & 7.8B \\
\bottomrule
\end{tabular}
}
\label{tab:model_config}
\end{table}

% \begin{tabularx}{\textwidth}{|X|X|X|X|X|X|X|X|}
% \hline
% Hidden Size & FFN Hidden Size & Layer Number & Heads & KV Heads & Sequence Length & Vocabulary Size & Parameters Number \\
% \hline
% 4096 & 13312 & 32 & 32 & 8 & 16384 & 152064 & 7.8B \\
% \hline
% \end{tabularx}

\subsection{Training Data Composition}
The training dataset for our table reasoning model is constructed from a heterogeneous collection of structured tables and free-form textual documents. 
The core content of table reasoning dataset is produced through our controlled data-generation pipeline with TCoT, PoT and ICoT reasoning traces, rather than sourced directly from raw corpora.
In contrast to generic LLM corpora, our dataset composition emphasizes reasoning tasks involving tabular data, numerical computation, and code generation.  
Prior to model training, all generated instances undergo an extensive cleaning and filtering process to ensure that the retained samples provide high-quality learning signals.

% Formally, we denote the initial raw dataset before any filtering as:
% \begin{equation}
%     \mathcal{D}_0 = \left\{ \big( \mathcal{T}_i, q_i, r_i, a_i \big) \right\}_{i=1}^N,
% \end{equation}
% where $\mathcal{T}_i = \{ t_{i,1}, t_{i,2}, \dots, t_{i,m_i} \}$ represents a collection of $m_i$ structured tables, 
% $q_i$ is the natural language question associated with these tables, $r_i$ is the reasoning trace (a step-by-step explanation) generated for answering the question, and $a_i$ is the final answer.  
% This formulation allows each sample to incorporate multi-table reasoning, ensuring coverage of complex table reasoning scenarios.

\subsubsection{Short Answer filtering}
In table reasoning, answers that are excessively short often fail to provide sufficient context or justification for model learning. 
Such instances risk training the model to memorize superficial input-output patterns instead of understanding the underlying reasoning process. 
Therefore, we introduce a filtering criterion that explicitly identifies and removes these cases based on answer length.
To formalize this filter, let $a_i$ denote the answer to sample $i$, $|a_i|_{\text{tok}}$ denote its token length, and $|a_i|_{\text{char}}$ denote its character length. We define the short-answer indicator as:
\begin{equation}
    f_{\text{short}}(i) = \mathbb{I}\big[ |a_i|_{\text{tok}} < \tau_{\text{tok}} \ \lor \ |a_i|_{\text{char}} < \tau_{\text{char}} \big],
\end{equation}
where $\tau_{\text{tok}}=5$ and $\tau_{\text{char}}=25$ are empirically chosen thresholds.  
Samples with $f_{\text{short}}(i)=1$ are discarded unless their shortness is accompanied by an explicit reasoning trace. This criterion ensures that LLMs can learn high-quality reasoning chain-of-thoughts from the dataset.

\subsubsection{Repetitive Thinking Filtering}
A significant issue in LLM-generated reasoning traces is the appearance of repeated substrings or redundant logical steps, which can reduce both conciseness and informativeness \cite{sui2025stop}. 
We detect such patterns using an improved rolling hash method \cite{karp1987efficient}, which computes a fingerprint for each fixed-size substring window and identifies repetitions beyond a set frequency.  
Let $r_i$ denote the reasoning trace for sample $i$, and let $\text{dup}(r_i)$ be the repeated times in $r_i$ using count of $n$-grams (with $n=10$ in our implementation). 
Then, the repetition indicator is defined as:
\begin{equation}
    f_{\text{repeat}}(i) = \mathbb{I}\big[ \text{dup}(r_i) > \tau_{\text{dup}} \big],
\end{equation}
where $\tau_{\text{dup}}=20$ is the maximum allowed repetition times. 
This step removes traces that loop over the same conclusion multiple times, enhancing concise reasoning of LLMs.

\subsubsection{Low Information Density Filtering}
Some reasoning traces, although lengthy, contain little substantive content because they are dominated by generic filler words. 
Such low-information samples introduce noise into the training process, particularly for table reasoning tasks where precision in referencing table elements is critical.  
Therefore, we adopt a filtering criterion based on a fixed \emph{low-information vocabulary} $\mathcal{V}_{\text{low}}$, which plays a role similar to a stop-word list. 
This vocabulary is curated to include high-frequency functional words, vague modifiers, and contrastive conjunctions that do not contribute meaningful semantic content to reasoning steps.

Formally, let $r_i$ be the reasoning trace of sample $i$, $T(r_i)$ the total number of tokens, and $L(r_i)$ the number of tokens from $\mathcal{V}_{\text{low}}$ appearing in $r_i$. We define the low-information density filter as:
\begin{equation}
    f_{\text{density}}(i) = \mathbb{I}\left[ \frac{L(r_i)}{T(r_i)} > \tau_{\text{low}} \right],
\end{equation}
where $\tau_{\text{low}}$ is a threshold controlling the maximum allowed proportion of low-information tokens (set to $0.5$ in our experiments).  
Samples with $f_{\text{density}}(i) = 1$ are discarded, ensuring that retained reasoning traces are semantically dense and contain sufficient content to support learning.

\subsubsection{LLM-based Scoring}
Rule-based heuristics can miss subtle forms of low quality, such as logically valid but incomplete answers. To address this, we introduce an LLM-based quality scoring stage.  
Let $s_i$ be the score assigned by a high-quality evaluator model (Qwen2-72B-Instruct in our work) to sample $i$, normalized to $[0,10]$ based on criteria including correctness, completeness, clarity and safety. The scoring-based filter is:
\begin{equation}
    f_{\text{score}}(i) = \mathbb{I}\big[ s_i < \tau_{\text{score}} \big],
\end{equation}
where $\tau_{\text{score}}=8.5$ is chosen to rigorously filter the samples. 
This step serves as a holistic safeguard, catching errors that escape other automated filters.

\subsubsection{Category Balancing}
After filtering, the number of training samples for different types of tasks may be unbalanced, especially between natrual language understanding and table reasoning. 
To ensure balanced training, we compute category proportions and perform targeted upsampling for underrepresented types.  
Formally, let $p_k$ be the proportion of category $k$ in the filtered dataset, and $p_{\text{target}}$ the desired uniform proportion. We adjust the sampling weights $w_k$ as:
\begin{equation}
    w_k = \frac{p^{\text{target}}_k}{p_k},
\end{equation}
which are applied to the filtered dataset to produce the final training dataset. 
This balancing prevents model overfitting to dominant categories and improves generalization.

\subsubsection{Overall Filtering Function}
Combining all filters, the final retained dataset $\mathcal{D}_{\text{final}}$ is obtained as:
\begin{equation}
    \mathcal{D}_{\text{final}} = \left\{ i \in \mathcal{D}_0 \ \middle| \ f_{\text{short}}(i) = 0 \ \land \ f_{\text{repeat}}(i) = 0 \ \land \ f_{\text{density}}(i) = 0 \ \land \ f_{\text{score}}(i) = 0 \right\},
\end{equation}
followed by category balancing weights $w_k$. 
$\mathcal{D}_0$ denotes the original raw dataset before any filtering.
The percentages of different task types are shown in Table~\ref{tab:category_percentage}.

\begin{table}[h]
\centering
\renewcommand{\arraystretch}{1.2} % 增加行高
\caption{Category distribution of training dataset of JT-DA-8B.}
\begin{tabular}{
    >{\raggedright\arraybackslash}p{6cm} % 左对齐列
    >{\centering\arraybackslash}p{3cm}   % 百分比列
}
\toprule
\textbf{Category} & \textbf{Percentage} \\
\midrule
Natural Language Understanding   & 23.65\% \\
Table Understanding              & 24.84\% \\
Table Basic Operation             & 31.16\% \\
Table Computational Operation     & 15.77\% \\
Data Analysis                     & 3.99\%  \\
Advanced Data Analysis            & 0.59\%  \\
\bottomrule
\end{tabular}
\label{tab:category_percentage}
\end{table}

\subsection{Training Details}
To improve the model’s performance on structured table reasoning tasks, we perform supervised fine-tuning (SFT) on a curated dataset emphasizing diverse reasoning patterns over tabular data.  
We adopt a teacher-forcing training paradigm, conditioning the model on the entire input context and training it to generate the target output sequence autoregressively, consistent with recent instruction tuning methodologies \cite{openai2023gpt4, qwen3technicalreport}.

Training is carried out on the Jiutian Intelligent Computing Platform, in which 96 NPUs interconnected via a high-bandwidth, low-latency network and a shared high-throughput storage system optimized for large-scale deep learning workloads are utilized.

We use the AdamW optimizer \cite{loshchilov2017decoupled} with an initial learning rate of $5 \times 10^{-6}$. 
To improve convergence, a cosine learning rate scheduler was applied, with a linear warm-up over the first 10\% of the total training steps.
The batch size is set to 96 samples per update step.

\section{Reinforcement Learning}
To further enhance the model’s capability, we introduce a reinforcement learning (RL) post-training mechanism to improve its ability to process tabular data.
% \subsection{Stage1}

\subsection{Datasets Preparation for RL}
In addition to the common data requirements for RL training, the dataset satisfies the following conditions:
 (1) not used during former training phases; (2) remains learnable for the model; (3) challenging; and (4) covers a broad range of sub-domains.
To address the limitations of existing open-source datasets for table-related training tasks—where answer reasoning is primarily performed in a TCoT mode, resulting in restricted table sizes and insufficient attention to table structure and content—this study proposes a reinforcement learning (RL) training framework based on table structures and partial table information. Using Program-of-Thought (PoT) reasoning, the model interacts with tabular data in a sandbox environment, reading CSV-format tables through environmental variables. This design ensures compatibility with both long and large tables.
The dataset consists of approximately 3,000 samples, ensuring both task diversity and an appropriate level of difficulty.

\subsection{Algorithm Design}
We adapt Group Relative Policy Optimization (GRPO) \cite{deepseekr1} including token-level loss computation. We omit KL penalty term present of original GRPO.
For each input $(q,a,t)$, the policy ${\pi_{\theta}}$ samples a group of ${G}$ candidate responses ${\{o_i\}_{i}^{G}}$.
Each response receives a reward $R_i$ from a judger system.
The group-normalized advantage for the $i-$th response at time step $t$ is:
\begin{equation}
    \hat{A}_{i,t}=\frac{R_i-mean(\{R_j\}_{j=1}^{G})}{std(\{R_j\}_{j=1}^{G})}.
\end{equation}

Our objective is optimized at the token-level with clipping:
\begin{equation}
    \mathcal{J}_{GRPO}(\theta)=\mathbb{E}_{(q,a)\sim{\mathcal{D},\{o_i\}_{i=1}^{G}\sim\pi_{\theta_{old}}(\cdot|q)}}[\frac{1}{\sum_{i=1}^{G}|o_i|}\sum_{i=1}^{G}\sum_{t=1}^{|o_i|}min(r_{i,t}(\theta)\hat{A}_{i,t},clip(r_{i,t}(\theta), 1-\epsilon_{low},1+\epsilon_{high})\hat{A}_{i,t})],
\end{equation}
the probability ratio $r_{i,t}(\theta)$ is defined as:
\begin{equation}
    r_{i,t}(\theta)=\frac{\pi_{\theta}(o_{i,t}|q,o_{i,<t})}{\pi_{\theta_{old}}(o_{i,t}|q,o_{i,<t})}.
\end{equation}

\subsection{Reward Design}
To achieve efficient table-task format training that is compatible with both wide and long tables, we design a reward framework based on PoT code execution accuracy, which consists of two components: result accuracy reward and format reward.
The result accuracy reward is provided by a verification system built upon PoT-based scoring, which includes an isolated sandbox and a result evaluator. The sandbox executes the table-analysis code generated by the model, while the result evaluator — implemented with a large language model to ensure compatibility with diverse QA pairs — scores the sandbox outputs against the ground-truth answers.
The format reward primarily parses the ${<think>}$...${</think>}$ field and assigns a base reward for correct format generation by the model.

% \subsection{Stage2}
% @zc @bs
\section{Table Reasoning Workflow Framework}
% Ce Chi, Ce Li, Kexin Yang, Chen Zhao, Boshen Shi

\begin{figure*}[tbp]
\centerline{\includegraphics[width=1.0\textwidth]{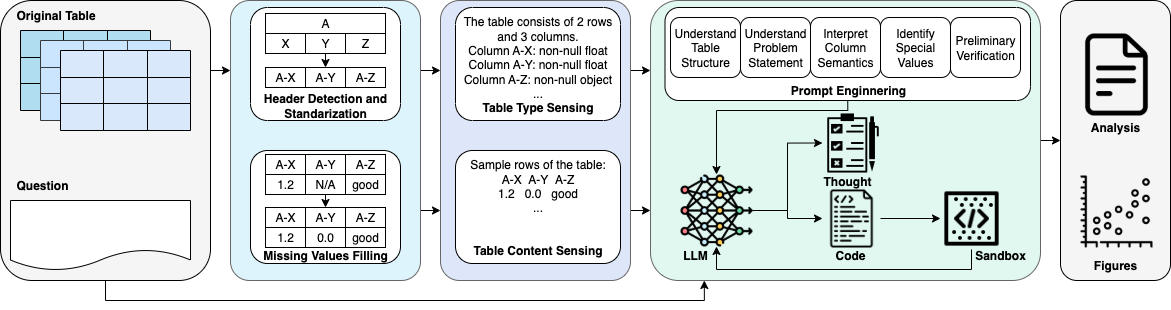}}
\caption{JT-DA workflow for table reasoning.}
\label{fig:workflow}
\end{figure*}

In order to fully leverage the reasoning capabilities of the trained model in tabular data scenarios, a structured data analysis-oriented workflow is proposed in this work. As illustrated in Fig~\ref{fig:workflow}, the workflow consists of 4 core components: table preprocessing, table sensing, tool-integrated reasoning and prompt engineering. These components form an end-to-end pipeline designed to enhance the model's ability to understand and reason over tabular datasets.

\subsection{Table Preprocessing}

Before table analysis and reasoning, a table preprocessing is adopted in our workflow such that the input table gets clean, structured and properly formatted. 
Table preprocessing involves handling missing values, splitting merged cells, identifying column headers, and standardizing column headers.
After the preprocessing, each table is transformed into a normalized structure such that downstream table understanding and reasoning can be facilitated. 
Moreover, the structured format reduces ambiguity in table layout and ensures consistent alignment between natural language queries and the corresponding data fields.

Formally, Given a collection of $N$ raw tables:
\[
\mathcal{T}^{raw} = \{ T^{raw}_1, T^{raw}_2, \dots, T^{raw}_N \},
\]
where each raw table \( T^{raw}_i \) contains a matrix of cells which can be splitted into table header and table body:
\[
T^{raw}_i = \{ c^{(i)}_{jk} \mid 1 \le j \le R_i, \ 1 \le k \le C_i \} = (H_i, B_i),
\]
where $R_i$ is the row number, $C_i$ is the column number, $H_i$ is the table header of the $i$-th table, and $B_i$ is its table body.
The preprocessing aims to produce a collection of $N$ processed tables:
\[
T^{proc}_i = \mathcal{P}(T^{raw}_i) = (H'_i, B'_i), \quad 1 \le i \le N,
\]

We define the overall preprocessing operator \(\mathcal{P}\) as the composition:
\[
\mathcal{P} = \big( \varphi_{\mathrm{std}} \times (\rho_{\mathrm{miss}} \circ \mu_{\mathrm{clean}}) \big) \circ \pi_{\mathrm{split}},
\]
where the operators act as follows on an input table \(T\):

\begin{align}
(H, B) &= \pi_{\mathrm{split}} (T), \label{eq:split} \\
H' &= \varphi_{\mathrm{std}}(H), \label{eq:header_std} \\
B' &= \rho_{\mathrm{miss}}(\mu_{\mathrm{clean}}(B)). \label{eq:body_clean}
\end{align}

\begin{itemize}
    \item \(\pi_{\mathrm{detect}}: \mathcal{T}_{R \times C} \to \mathcal{T}_{(R-R') \times C} \times \mathcal{T}_{R' \times C}\) detects and splits the table into header \(H\) and body \(B\).
    \item \(\varphi_{\mathrm{std}}: \mathcal{T}_{(R-R') \times C} \to \mathcal{T}_{1 \times C}\) standardizes the header rows.
    \item \(\mu_{\mathrm{clean}}: \mathcal{T}_{R' \times C} \to \mathcal{T}_{R' \times C}\) further processes the body \(B\), e.g., splitting merged cells.
    \item \(\rho_{\mathrm{miss}}: \mathcal{T}_{R' \times C} \to \mathcal{T}_{R' \times C}\) resolves missing or malformed values in the raw table.
\end{itemize}

After the preprocessing, the tables can be represented as:
\[
\mathcal{T}^{proc} = \{ T_i^{proc} \mid 1 \le i \le N \}.
\]

\subsection{Table Sensing}

Table sensing refers to the model’s contextual understanding of the tables’ structural and basic statistical information. 
In this stage, basic observations such as column headers, data types, and representative sample rows of each table are collected and presented to the model. 
This provides the model with a concise overview of the table’s schema and content distribution, facilitating informed interpretation.

The sensing stage extracts a compact summary representation capturing key structural and statistical properties. 
Formally, given a processed table
\[
T^{proc}_i = (H'_i, B'_i).
\]
We define the sensing function
\[
\mathcal{S} : \mathcal{T} \to \mathcal{O},
\]
mapping a processed table \(T^{proc}_i\) to metadata \(\mathbf{O}_i \in \mathcal{O}\), where
\[
\mathbf{O}_i = \big( H'_i,  \mathbf{t}_i,  \mathbf{m}_i,  (R_i, C_i),  B''_i \big),
\]
with
\begin{itemize}
    \item \(H'_i\): the set of column headers.
    \item \(\mathbf{t}_i = (t_1^{(i)}, t_2^{(i)}, \ldots, t_{C_i}^{(i)})\): inferred data types for each column (e.g., numerical, categorical, textual).
    \item \(\mathbf{m}_i = (m_1^{(i)}, m_2^{(i)}, \ldots, m_{C_k}^{(i)})\): missing value counts per column,
    \item \((R_i, C_i)\): the table dimensions, i.e., number of rows \(R_i\) and columns \(C_i\),
    % \item \(B''_i \subseteq B'_i\): a sample subset of rows from the body \(B'_i\), used when \(R_i\) is large to reduce context size.
    \item \(B''_i \subseteq B'_i\): a sample subset of rows from the body \(B'_i\), utilized to reduce the context size when \(R_i\) is large.
\end{itemize}

This metadata \(\mathbf{O}_i\) summarizes the table’s structural schema and key statistics, serving as the primary input for downstream models or reasoning systems. Notably, missing value counts \(\mathbf{m}_i\) are retained in case that the missing values filling fails in the preprocessing stage. And the sample \(B''_i\) preserves representative content without overwhelming model context windows.
Such streamlined table sensing strikes a balance between rich semantic understanding and practical resource constraints, making it well-suited for large-scale table reasoning pipelines.

\subsection{Tool-Integrated Reasoning}

To simulate the interactive process between humans and structured data, our system integrates multiple external tools, including visualization module $\mathcal{I}_{vis}$ and code sandboxes $\mathcal{I}_{code}$, into our iterative table reasoning loop. 
% The integration of tools can effectively alleviate the hallucination problem of LLM in numerical calculations. 
These tools enable the model to offload precise computation, data manipulation, and visualization tasks, thereby reducing hallucination, improving factuality, and enhancing interpretability.
Moreover, by interacting with the external tools, the model can incrementally refines its internal representation and understanding of the table so as to improve the overall performance of table reasoning. Finally, the interaction between LLMs and sandboxes also improves the model's interpretability and reliability. 

Formally, let the input query be \(q\) and the table sensing metadata extracted from the processed tables be \(\ \{\mathbf{O}_i\}_{i=1}^N\). The combined input at the start is
\[
x = (q, \{\mathbf{O}_i\}_{i=1}^N),
\]
the reasoning process proceeds over \(K\) steps as defined as following iterative steps:
\[
\begin{aligned}
& o_k = \mathcal{M}\big(x, h_{k-1}, r_{k-1}\big), \\
& r_k = \mathcal{I}(o_k), \\
& h_k = \big(h_{k-1}, o_k, r_{k-1}\big).
\end{aligned}
\]

with initial conditions \(h_0 = \varnothing\) and \(r_0 = \varnothing\).

Here,
\begin{itemize}
    \item \(h_k\) denotes the model’s internal state at step \(k\), which cumulatively aggregates reasoning context and history of prior interactions.
    \item \(o_k\) is the model’s output at step \(k\), including instructions or code snippets for the tools.
    \item \(r_k\) is the tool’s execution result returned to the model at step \(k\).
    \item \(\mathcal{M}(\cdot)\) is the model inference function conditioning on the combined input \(x\), previous internal state \(h_{k-1}\), and last tool output \(r_{k-1}\).
    \item $\mathcal{I}\in\{\mathcal{I}_{vis}, \mathcal{I}_{code}\}$ represents a unified tool interface, under which both the code sandbox and ChartTool are specific realizations.
\end{itemize}

By delegating precise numerical calculation and data manipulation to the tools \(\mathcal{T}\), this approach effectively curbs hallucination errors and provides transparent intermediate results, enhancing interpretability and trustworthiness \cite{yang2025code}.

\textbf{Vision Tool Calling}. To enable LLMs to perform complex data visualization beyond text generation, we designed and implemented a function-call–based visualization tool into our tool-augmented reasoning framework. 
This module, referred to as \textbf{ChartTool}, enables the model to generate structured and executable data visualizations without requiring explicit knowledge of underlying rendering libraries. 
This approach not only expands the application scenarios for LLMs but also ensures code execution security and stability through a robust abstraction layer.

Given a natural-language instruction 
$q$, the LLM produces a structured function call
\begin{equation}
    c = \Phi(q),
\end{equation}
where $\Phi$ maps the instruction to ChatTool's parameter space via LLMs.
Afterwards, ChatTool $\mathcal{I}_{vis}$ transforms the function call into a rendered visualization:
\begin{equation}
    v = \mathcal{I}_{vis}(c),
\end{equation}
where $\mathcal{I}_{vis}$ denotes the visualization pileline of ChatTool.

Beyond its role as a lightweight visualization backend, ChartTool serves as a crucial component in enabling reliable data visualization.
By enforcing a unified parameter schema, it standardizes heterogeneous visualization tasks into a consistent function-call protocol, allowing the model to generalize across chart types and configurations with minimal supervision.
Moreover, this structured design not only improves execution correctness but also significantly enhances the interpretability of the model's reasoning process.

\subsection{Prompt Engineering}

Prompt engineering plays a critical role in guiding the model through multi-step table reasoning processes. To address the inherent challenges in tabular data understanding and reasoning, we propose a prompt optimization strategy specifically designed for table analysis. This strategy focuses on enhancing the model’s contextual understanding and reasoning accuracy by decomposing the task into interpretable components. The core objectives of the prompt design are as follows:

\textbf{Understanding Table Structure}.
The model should recognize the structural properties of the table. This includes identifying multi-row headers, detecting summary rows or columns (e.g., totals or averages), and flagging potentially noisy or dirty data (e.g., non-standard entries in the last row or column). The model is also expected to distinguish between data-bearing and metadata-bearing rows or columns.

\textbf{Understanding the Problem Statement}.
For reasoning tasks where the query is abstract, ambiguous, or underspecified, the model should deconstruct the problem into more concrete, answerable sub-questions. This step helps bridge the gap between vague instructions and the structured logic required for accurate table interpretation.

\textbf{Interpreting Column Semantics}.
The model should infer the semantic role of each column in relation to the question. For instance, distinguishing whether a column contains raw values, ratios, percentages, categories or derived metrics is critical for selecting the correct reasoning approach. 

\textbf{Identifying Special Values and Formatting Artifacts}.
Tabular data often contains non-standard values, such as null entries, special symbols, unit annotations or formatting elements like commas and percentage signs. The model must handle such artifacts correctly—either by preprocessing, filtering or normalization—to avoid misinterpretation during reasoning. 

\textbf{Preliminary Verification of Results}.
To further enhance reliability, the model is instructed to perform preliminary calculations before committing to a final answer. For example, if a proportion exceeds 100\%, it may indicate a misinterpretation of the table or the problem. This verification step serves as a safeguard against misinterpretation of the table or misalignment with the problem context.

\tcbset{colback=gray!10, colframe=gray!80, boxrule=0.5pt, arc=4pt}

\begin{tcolorbox}[title=Prompt Engineering]
A good data analysis process may include:
\vspace{1em}

\#\#\# Understanding stage

1. Understand the table structure, such as the column name may have two rows, the last row is the summary, the last column is the summary, the last row is the dirty data, etc.

2. Understand the problem, such as when the problem is more abstract and vague, should be analyzed into a specific problems.

3. Understand the column meaning related to the problem, such as whether the column name represents the actual value or the ratio.

4. Identify special values in the table, such as columns and rows may have null values, special symbols, commas, etc.

\vspace{1em}
\#\#\# Solution stage

1. Sort out the solution ideas, such as how to clean the data and how to write the solution code.

2. Calculate the result by hand to judge whether the idea is correct, such as the proportion should not exceed 100\%, otherwise the table understanding or problem understanding may be not in place.

3. Summarize the processing process and precautions, which is convenient for reference when writing code later.
\end{tcolorbox}

\subsection{Overall Workflow Formalization}

Given a set of raw tables \(\mathcal{T}^{raw} = \{T_1^{raw}, \ldots, T_N^{raw}\}\) and a natural language query \(q\), the final answer of the table reasoning process can be formalized as 
\[
\begin{aligned}
    y &= \mathrm{Decode}(h_K),
\end{aligned}
\]

where $h_K$ is iteratively calculated by
\[
\begin{cases}
x = \big(Prompt, q, \{\mathbf{O}_i\}_{i=1}^N \big), \\
\mathbf{O}_i = \mathcal{S}(\mathcal{P}(T_i^{raw})), \\
o_k = \mathcal{M}\big(x, h_{k-1}, r_{k-1}\big), \\
r_k = \mathcal{I}(o_k), \\
h_k = f(h_{k-1}, o_k, r_{k-1}),
\end{cases}
\quad k=1, \ldots, K,
\quad h_0 = \varnothing, \quad r_0 = \varnothing.
\]

$\mathrm{Decode}(\cdot)$ denotes an answer extracting process from LLM outputs.

This unified workflow provides a principled formulation of tool-augmented table reasoning, where perception, planning, visualization, execution, and verification are tightly integrated into a coherent iterative loop.
The framework ensures that each reasoning step is both interpretable and operationally grounded.
Moreover, the synergy between table preprocessing, model-generated instructions, and executable feedback enables the workflow to progressively refine its internal representation and converge toward reliable solutions, even for complex analytical tasks.
\section{Evaluation}

\subsection{Experiment Setup}

\subsubsection{Open Benchmark}
The model evaluation is based on an open-source benchmark: TReB~\cite{li2025treb}. It is a reliable and comprehensive benchmark that offers assessment of LLM capabilities for table reasoning. 

The benchmark provides an open-source dataset combining cleaned public benchmarks, real-world web tables, and proprietary data to support diverse evaluations. The testset is divided into five table reasoning aspects:

\begin{itemize}
\item \textbf{Table Understanding (TU)}: Assesses the ability to parse table structures and comprehend full or partial table content across 6 subtasks.

\item \textbf{Table Basic Operation (TBO)}: Measures the accuracy of mapping natural language intents to fundamental table operations , with 2 subtasks.

\item \textbf{Table Computational Operation (TCO)}: Tests the ability to execute complex computational procedures in table reasoning scenarios, through 2 subtasks.

\item \textbf{Data Analysis (DA)}: Focuses on statistical analysis and pattern recognition in table reasoning scenarios across 4 subtasks.

\item \textbf{Advanced Data Analysis (ADA)}: Targets multi-step ($\geq$ 3 steps) analytical reasoning across 6 subtasks.
\end{itemize}

Additionally, the benchmark provides an open-source framework code specifically designed to evaluate LLM performance on table reasoning tasks. It integrates diverse inference modes and reliable metrics, enabling precise and multi-dimensional evaluations.
Specifically, 3 inference modes are provided:

\begin{itemize}
\item \textbf{Textual Chain-of-Thought (TCoT)}: LLMs solve problems step by step through pure textual reasoning. The final answer is output exclusively in text form.

\item \textbf{Program-of-Thought (PoT)}: LLMs solve problems by generating executable code. 

\item \textbf{Interleaved Chain-of-Thought (ICoT)}: This mode enables models to perform multi-step reasoning by combining textual explanations and programmatic outputs in an iterative process.
\end{itemize}

In this work, we use LLM-as-a-judge as evaluation metrics, while maintaining the same experimental settings as TReB~\cite{li2025treb}.

\subsubsection{Baseline models}
We set a diverse range of models designed for different purposes as the baselines. The evaluation model size is selected to be approximately the same as JT-DA-8B, ensuring that the model is assessed under comparable scaling laws.

\textbf{General LLMs}: The general LLMs, which represent the baseline performance of language models on table reasoning, include Llama-3.1-8B-Instruct~\cite{dubey2024llama}, Qwen2.5-7B-Instruct~\cite{qwen2.5}, and Mistral-7B-Instruct-v0.3~\cite{jiang2023mistral}.

\textbf{Code Optimized LLMs}: The code-optimized LLMs, trained with a focus on code generation, are evaluated to explore their potential in handling table reasoning. This group includes Qwen2.5-Coder-7B-Instruct~\cite{hui2024qwen2}, Seed-Coder-8B-Instruct, and Yi-Coder-9B-Chat~\cite{ai2024yi}. 

\textbf{Deep Thinking LLMs}: Deep thinking LLMs are designed to excel in complex problem analysis and self-reflective reasoning. This group includes DeepSeek-distilled variants of Qwen-7B and Llama-8B~\cite{deepseekr1}.

\textbf{Table Reasoning Optimized LLMs}: We incorporate three specific LLMs, TableGPT2-7B~\cite{su2024tablegpt2} and Table-R1-SFT/Zero-7B~\cite{yang2025table}, which are fine-tuned specifically for the table reasoning. 

\begin{table*}[htbp]
  \centering
  \caption{Overall experimental results with LLM-as-a-judge.}
  \resizebox{\textwidth}{!}{
    \begin{tabular}{lcccccccccccccccr}
    \toprule
    \multicolumn{1}{c}{\multirow{2}[4]{*}{\textbf{Model Name}}} & \multicolumn{3}{c}{\textbf{TU}} & \multicolumn{3}{c}{\textbf{TBO}} & \multicolumn{3}{c}{\textbf{TCO}} & \multicolumn{3}{c}{\textbf{DA}} & \multicolumn{3}{c}{\textbf{ADA}} & \multicolumn{1}{c}{\multirow{2}[4]{*}{\textbf{Overall}}} \\
\cmidrule(lr){2-4} \cmidrule(lr){5-7} \cmidrule(lr){8-10} \cmidrule(lr){11-13} \cmidrule(lr){14-16}          & \multicolumn{1}{c}{\textbf{TCoT}} & \multicolumn{1}{c}{\textbf{PoT}} & \multicolumn{1}{c}{\textbf{ICoT}} & \multicolumn{1}{c}{\textbf{TCoT}} & \multicolumn{1}{c}{\textbf{PoT}} & \multicolumn{1}{c}{\textbf{ICoT}} & \multicolumn{1}{c}{\textbf{TCoT}} & \multicolumn{1}{c}{\textbf{PoT}} & \multicolumn{1}{c}{\textbf{ICoT}} & \multicolumn{1}{c}{\textbf{TCoT}} & \multicolumn{1}{c}{\textbf{PoT}} & \multicolumn{1}{c}{\textbf{ICoT}} & \multicolumn{1}{c}{\textbf{TCoT}} & \multicolumn{1}{c}{\textbf{PoT}} & \multicolumn{1}{c}{\textbf{ICoT}} &  \\
    \midrule
    \multicolumn{17}{c}{\textit{\textbf{General LLMs}}} \\
    \midrule
    Llama-3.1-8B-Instruct & 49.53  & 45.04  & 47.06  & 39.22  & 50.11  & 55.76  & 41.51  & 55.12  & 50.20  & 39.68  & 51.97  & 50.89  & 31.36  & 24.01  & 13.62  & 43.00  \\
    Qwen2.5-7B-Instruct & 58.79  & 57.98  & 68.26  & 41.46  & 56.99  & 61.23  & 51.61  & 62.71  & 73.63  & 44.88  & 52.17  & 61.84  & 37.13  & 28.08  & 44.63  & 53.42  \\
    Mistral-7B-Instruct-v0.3 & 40.22  & 37.33  & 25.14  & 32.57  & 50.09  & 37.18  & 35.28  & 45.46  & 35.61  & 37.20  & 23.96  & 27.09  & 18.89  & 15.14  & 16.05  & 31.81  \\
    \midrule
    \multicolumn{17}{c}{\textit{\textbf{Code Optimized LLMs}}} \\
    \midrule
    Qwen2.5-Coder-7B-Instruct & 55.51  & 57.92  & 64.30  & 44.36  & 61.03  & 66.86  & 47.59  & 65.17  & 69.26  & 39.99  & 57.73  & 62.99  & 35.99  & 33.97  & 45.50  & 53.88  \\
    Seed-Coder-8B-Instruct & 50.65  & 57.21  & 59.61  & 44.06  & 66.58  & 66.33  & 42.66  & 67.57  & 68.67  & 38.68  & 62.94  & 65.62  & 32.59  & 38.05  & 42.27  & 53.57  \\
    Yi-Coder-9B-Chat & 32.76  & 50.61  & 39.70  & 30.59  & 56.24  & 51.56  & 32.49  & 58.34  & 54.58  & 23.07  & 48.17  & 51.50  & 21.41  & 28.72  & 20.62  & 40.02  \\
    \midrule
    \multicolumn{17}{c}{\textit{\textbf{Deep Thinking LLMs}}} \\
    \midrule
    Deepseek-R1-Distill-Qwen-7B & 49.28  & 33.86  & 54.58  & 56.33  & 48.24  & 52.49  & 62.01  & 43.28  & 56.50  & 49.19  & 41.19  & 57.34  & 22.86  & 18.02  & 36.89  & 45.47  \\
    Deepseek-R1-Distill-Llama-8B & 56.13  & 34.86  & 54.48  & 55.23  & 39.07  & 49.60  & 57.29  & 36.86  & 52.17  & 49.24  & 13.14  & 40.90  & 20.32  & 6.40  & 25.19  & 39.39  \\
    \midrule
    \multicolumn{17}{c}{\textit{\textbf{Table Reasoning Optimized LLMs}}} \\
    \midrule
    TableGPT2-7B & 58.97  & 64.61  & 73.82  & 48.38  & 59.79  & 65.43  & 57.05  & 71.94  & 75.58  & 44.14  & 51.24  & 66.80  & 32.06  & 33.21  & 50.68  & 56.91  \\
    TableR1-SFT-7B & 62.04  & 53.52  & 25.10  & 68.21  & 54.49  & 15.70  & \textbf{71.25}  & 59.87  & 13.57  & 37.08  & 41.26  & 33.75  & 35.64  & 25.95  & 19.95  & 41.16  \\
    TableR1-Zero-7B & 64.36  & 48.99  & 77.99  & 54.56  & 36.01  & 58.57  & 62.06  & 50.28  & 76.16  & 50.32  & 48.95  & 63.79  & 28.21  & 32.32  & 45.36  & 53.19  \\
    \textbf{JT-DA-8B} & 59.97 & \textbf{76.02} 	& 75.37 	& 44.79 	& \textbf{91.84} 	& \textbf{87.62} 	& 41.17 	& 74.99 	& 77.72 	& 47.47 	& \textbf{74.13} 	& \textbf{80.34} 	& 39.30 	& 35.56 	& \textbf{51.65} & 63.86 \\
    \textbf{JT-DA-R1-8B} & \textbf{65.44} & 73.28 & \textbf{78.88} & \textbf{80.02} & 87.84 & 85.01 & 63.93 & \textbf{76.10} & \textbf{78.83} & \textbf{53.59} & 72.17 & 76.59 & \textbf{42.10} & \textbf{39.96} &	49.06 & \textbf{68.18}

   \\
    \bottomrule
    \end{tabular}%
    }
  \label{tab:res-llm-score}%
\end{table*}%

\subsection{Experimental Result}

\subsubsection{Model Performance Study}
The average performance of all models across all tasks is shown in Table \ref{tab:res-llm-score}. The experimental results reveal several key observations regarding the performance of the JT-DA. Notably, JT-DA demonstrated outstanding performance across a wide range of table reasoning tasks, achieving the highest overall score among models of a comparable scale.

JT-DA excelled in tasks involving TBO, which primarily include precise and fuzzy data queries. In scenarios where code execution was enabled, the model achieved table query scores exceeding 90, indicating exceptional accuracy in such tasks. This precise table querying ability forms a strong foundation for the model's overall performance, contributing significantly to its capabilities in table computation and data analysis. However, it is evident that the model exhibits certain weaknesses in TCoT reasoning. This limitation may stem from the model's training emphasis on leveraging chain-of-thought and code execution specifically for table reasoning tasks, potentially leading to a partial forgetting of broader textual reasoning diversity.

Furthermore, in ADA tasks requiring multi-step reasoning, JT-DA demonstrated superior performance in the ICoT mode compared to other models of similar scale. This suggests that the model is well-suited for real-world scenarios involving multi-round, interactive reasoning, where the ability to handle long reasoning chains and solve complex data analysis problems is critical. These results underscore the model's robustness and scalability in addressing sophisticated reasoning tasks within tabular data contexts.

Through on-policy training with the GRPO algorithm, the JT-DA-R1 model achieved a significant performance improvement compared to the SFT model. Specifically, the overall score of JT-DA-R1 increased by 4.5\% relative to the SFT baseline. This improvement is primarily attributed to enhanced reasoning capabilities in the TCoT mode. Across five evaluation tasks in TCoT, JT-DA-R1 outperformed the SFT model by 5.47\%, 35.23\%, 22.76\%, 6.12\%, and 2.8\%, respectively. Additionly, the model’s performance in the PoT and ICoT modes remained nearly unchanged. These results confirm that reinforcement learning can effectively enhance the table reasoning abilities of the model.

\subsubsection{Case Study}
In this section, we present a series of cases to demonstrate the powerful data analysis capabilities of JT-DA-8B.

\begin{figure*}[tbp]
\centerline{\includegraphics[width=0.8\textwidth]{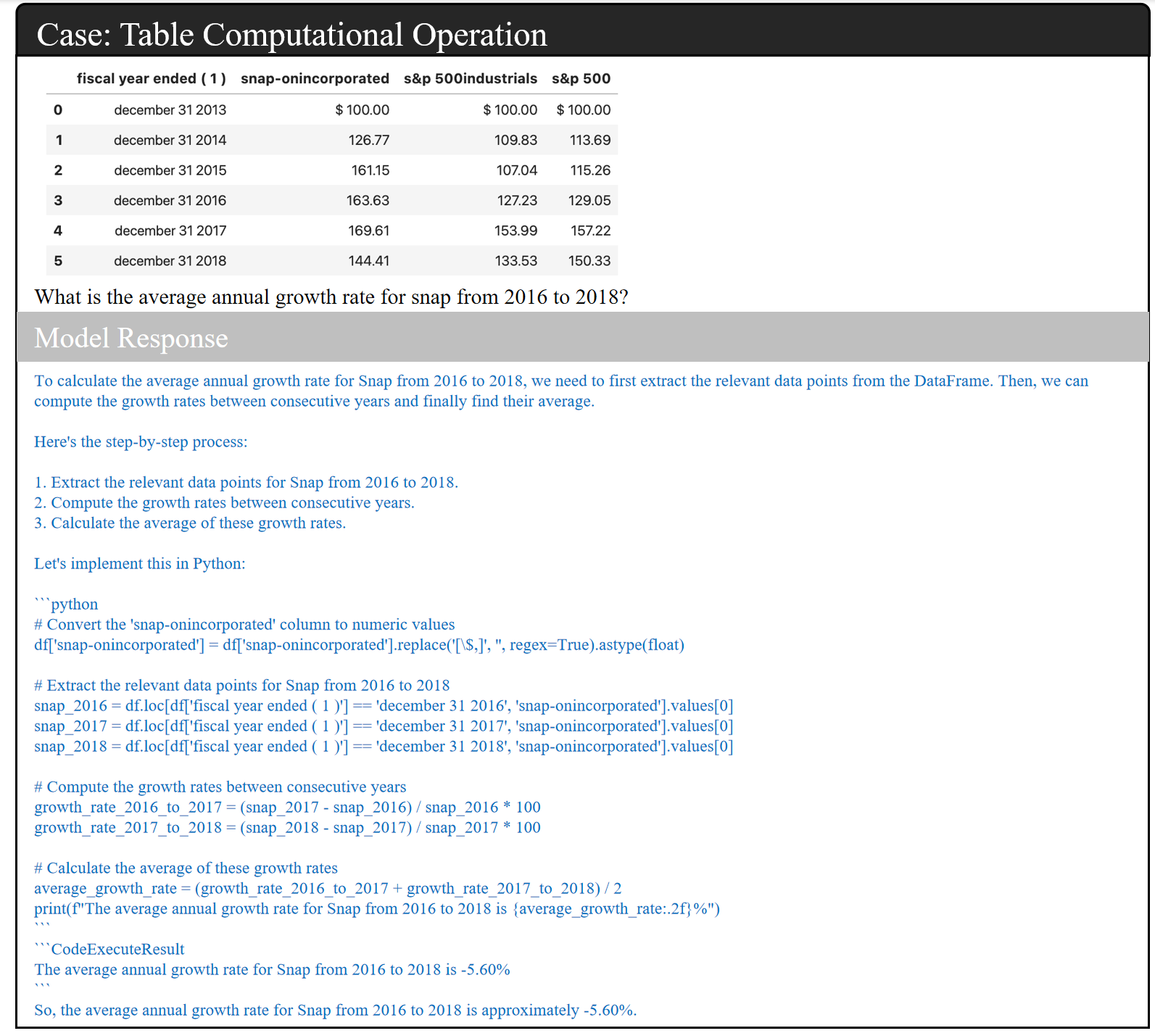}}
\caption{Case1: Table computational operation.}
\label{fig:case1}
\end{figure*}

\begin{figure*}[htbp]
\centerline{\includegraphics[width=0.8\textwidth]{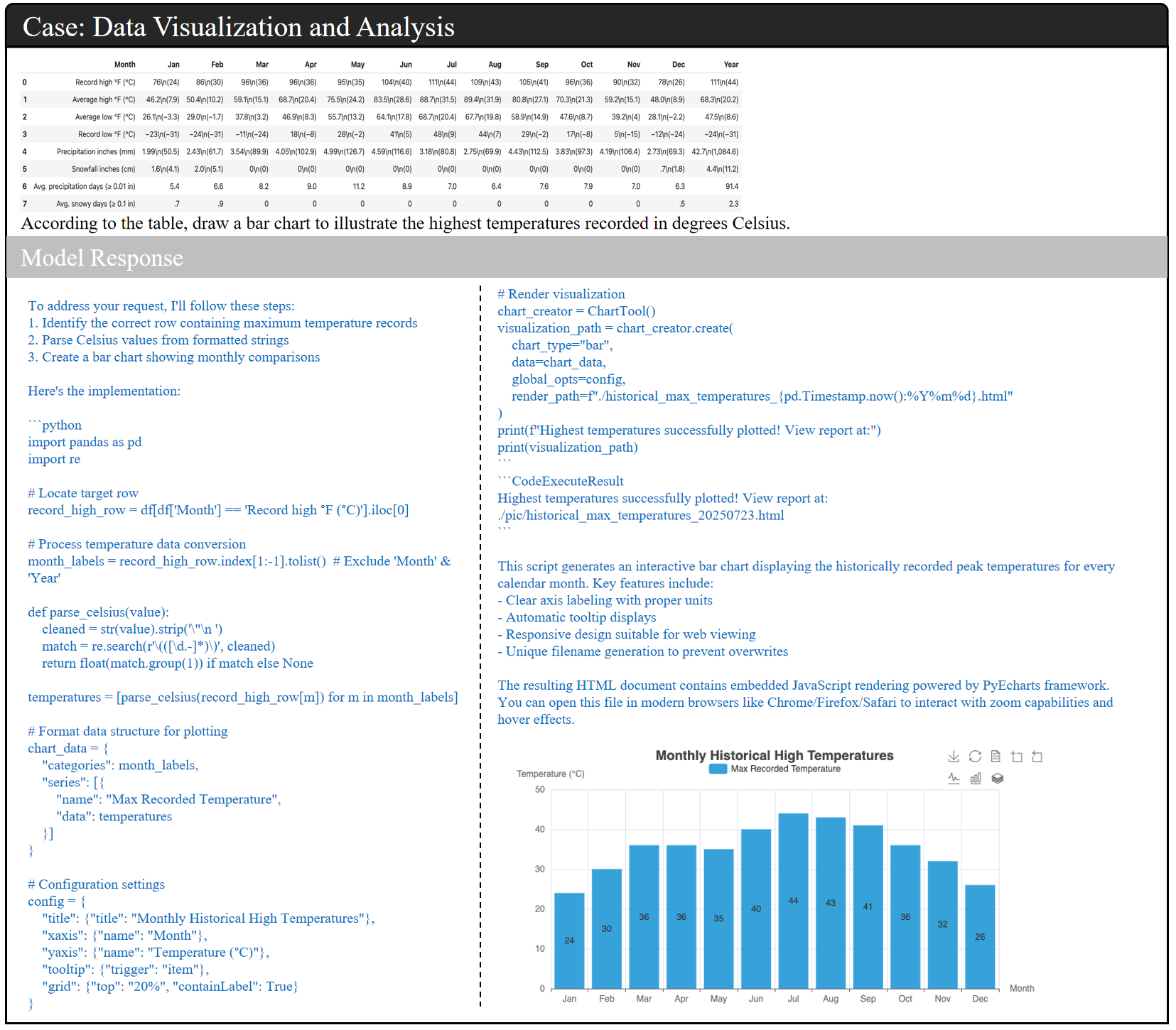}}
\caption{Case2: Data visualization.}
\label{fig:case2}
\end{figure*}

Case 1, illustrated in Figure \ref{fig:case1}, provides a comprehensive evaluation of the model's capabilities in addressing table reasoning tasks. Specifically, this case demonstrates the model's proficiency in three critical areas: table and question understanding, data query pre-screening, and accurate data computation. The problem requires the model to extract relevant data points from a structured yet noisy dataset, compute growth rates between consecutive years, and then derive their average to answer the question regarding Snap’s average annual growth rate from 2016 to 2018. 

A key challenge in Case 1 lies in the dataset's inconsistent and noisy data types, such as monetary values stored as strings with symbols (e.g., "\$"). The model successfully handles this issue by employing pre-processing steps, including the conversion of string data into numeric formats. This demonstrates its ability to adapt to real-world datasets where such inconsistencies are common. Furthermore, the model showcases a clear understanding of the question by breaking down the task into logical steps: data extraction, computation of year-on-year growth rates, and their averaging. This structured approach highlights the model's strength in reasoning and decomposition of complex computational tasks. The final output, an average annual growth rate of -5.60\%, is both mathematically accurate and contextually appropriate, reflecting the model's effectiveness in combining accurate computations with robust data querying capabilities. This case underscores the model's ability to process tabular data efficiently, even when faced with noisy or inconsistent formats, and to generate precise and interpretable results. Such performance demonstrates the model's suitability for real-world applications in table reasoning and data analysis tasks.

Case 2, illustrated in Figure \ref{fig:case2}, evaluates the model's capability in data visualization and analysis, particularly in the context of tabular reasoning tasks. The problem requires the model to generate a bar chart illustrating the highest temperatures recorded in degrees Celsius for each calendar month based on the provided table.

The model demonstrates a robust understanding of the task by following a structured workflow, including identifying the correct row in the dataset, parsing temperature values in Celsius from formatted strings, and configuring the chart with appropriate labels and data points. Parsing noisy and irregular data formats, such as extracting Celsius values from mixed strings containing Fahrenheit and Celsius in parentheses, highlights the model's ability to handle real-world datasets with inconsistent formatting.

The resulting bar chart exhibits several desirable qualities for effective data visualization. These include clearly labeled axes with proper units, an interactive and visually appealing representation of the data, etc, making it suitable for exploratory data analysis and presentation in professional or academic settings.

\section{Related Work}
\subsection{LLMs for Data Analysis}
\subsubsection{LLMs for General Data Analysis}
The application of LLMs in the field of data analysis is rapidly evolving from proof-of-concept studies to practical, real-world deployments~\cite{cheng2023gpt}. Depending on specific tasks, LLMs should be capable of understanding data, manipulating data, and conducting downstream tasks like prediction and classification. Recently, numerous studies have advanced such capabilities of LLMs, including data cleaning and pre-processing~\cite{zhang2023large,zhang2024jellyfish,yan2024efficient,pandas-ai,jupyter-ai}, data exploration (EDA)~\cite{ma2023demonstration,wang2025jupybara,han2024large,EDA-GPT}, anomaly detection~\cite{xu2024large,han2023loggpt}, visualization~\cite{goswami2025plotgen,yang2024matplotagent}, and solving prediction or classification tasks~\cite{zhang2023mlcopilot,jing2024dsbench,trirat2024automl}. In real-world applications, data-analysis LLMs have been applied to domains such as finance, medical and health care, E-commerce~\cite{liu2025findabench,singhal2025toward,wang2024enabling}. 

\subsubsection{LLMs for Tabular Reasoning}
Tables serve as a structured and interpretable format for representing complex data, playing a central role in a wide range of data analysis tasks. To perform downstream operations such as information retrieval, numerical computation, or trend forecasting, LLMs must possess robust table understanding and reasoning capabilities~\cite{li2025treb}. This underscores the importance of developing table-specific modeling approaches tailored to the unique challenges of tabular data.

While early research primarily focused on pretraining smaller language models for specialized table tasks~\cite{dong2022table,zhao2022reastap,kim2025table}, recent advances have shifted toward adapting LLMs to generalize across diverse table reasoning tasks through techniques such as supervised fine-tuning (SFT) and reinforcement learning (RL)~\cite{zhang2024tablellm,su2024tablegpt2,zhang2023tablellama,yang2025table,wu2025table,jin2025table}. Given the inherent complexity of table mining, which often involves multiple interdependent subtasks including retrieval, numerical computation, and logical reasoning, various inference-time scaling methods have been proposed to enhance model performance. Two prominent directions include complex reasoning frameworks and agentic workflows. The former aims to guide LLMs through structured reasoning steps tailored to tabular inputs~\cite{chen2024tablerag,ye2023large,wang2024chain}, while the latter introduces mechanisms such as memory and reflection to support more sophisticated and iterative reasoning processes~\cite{yu2025table}. 

Although some progress has been made, there are still two challenges that need to be addressed. Firstly, current table LLMs have a relatively small number of parameters(most of them are post-trained on 7B models), and the number of datasets used for training is small, with average quality and a lack of complex data. In addition, the data analysis capability of LLM itself should be enhanced, instead of always relying on the agentic framework for iterative trial-and-error and correction. Therefore, we train and release \modelname, which inherently endows with strong data analysis capabilities.

\subsection{Reasoning Capabilities of LLMs}
Research on enhancing the reasoning capabilities of LLMs began with the foundational Chain-of-Thought (CoT) prompting, which elicits intermediate steps to solve complex problems. Following CoT~\cite{wei2022chain}, researchers have proposed numerous improvements, including Symbolic Chain-of-Thought~\cite{li2023symbolic}, Chain-of-Abstraction~\cite{gao2024efficient}, Chain-of-Logic~\cite{servantez2024chain}, Tree-of-Thoughts~\cite{yao2023tree}, Graph-of-Thoughts~\cite{besta2024graph}, and many other prompting strategies based on ReAct~\cite{yao2023react} framework such as Program-of-Thoughts~\cite{chen2022program}. Considering the complex nature of data analysis tasks, which often requires models to invoke tools, execute code, or engage in self-reflection to arrive at accurate conclusions, we propose an ICoT (Interleaving CoT) mode. It enables models to perform multi-step reasoning by combining textual explanations and programmatic outputs in an iterative process.

Building upon the CoT paradigm, researchers have proposed distillation techniques aimed at mitigating the reliance of CoT methods on large scale of parameters, thereby enabling smaller models to achieve comparable reasoning capabilities. The most popular and core mechanism involves post-training a compact student model with fixed datasets containing reasoning traces from larger teacher models~\cite{guha2025openthoughts,openr1,muennighoffs1,sky_t1_2025,deepseekr1} (also referred to as `off-policy distillation'~\cite{qwen3technicalreport}). In \modelname, we also adopt such `off-policy distillation' during SFT.

% 加一个TIR的相关工作
\section{Conclusion and Future Work}
In this work, we presented JT-DA, a tool-integrated LLM with workflows designed to address the structural awareness and symbolic reasoning limitations of existing LLMs in table reasoning tasks.
By introducing an efficient data synthesization pipeline, we generated millions of high-quality and diverse training examples spanning multiple table reasoning categories.
Through large-scale SFT and RL, our model internalizes tool-augmented reasoning patterns, enabling it to more accurately generate its reasoning process in table structures and to dynamically integrate external computation tools during inference.
Extensive evaluations across multiple table reasoning tasks demonstrated that JT-DA consistently outperforms existing state-of-the-art methods in both accuracy and interpretability.

Despite the advances, JT-DA still has limitations. 
First, its current workflow is static, which can restrict adaptability to diverse analytical tasks. 
Second, NL2SQL capabilities is not trained and integrated, limiting direct database interaction. 
Third, the model’s visualization outputs are functional but not yet rich enough for interactive exploration. 

To overcome these challenges, our future work will focus on:
\begin{enumerate}
    \item \textbf{Self-improving table agents}: Enabling JT-DA to autonomously generate, reason, and solve more complex table reasoning tasks such as data analysis report generation.
    \item \textbf{NL2SQL integration}: Developing robust natural language to SQL translation over arbitrary schemas, enabling direct, reliable querying of large-scale databases.
    \item \textbf{Enhanced visualization}: Producing rich, stable charts, annotated tables, and step-by-step visualization views to improve data analysis experience and insights.
    \item \textbf{Optimized table encoding methods}: Designing more expressive and schema-aware table representation techniques for two-dimensional tables to improve reasoning accuracy of LLMs.
\end{enumerate}

\bibliography{main}
\bibliographystyle{colm2024_conference}

\end{document}